\documentclass{article} 
\usepackage{iclr2025_conference,times}

\usepackage{amsmath,amsfonts,bm}

\def\eqref#1{equation~\ref{#1}}

\def\1{\bm{1}}

\DeclareMathAlphabet{\mathsfit}{\encodingdefault}{\sfdefault}{m}{sl}
\SetMathAlphabet{\mathsfit}{bold}{\encodingdefault}{\sfdefault}{bx}{n}

\usepackage{hyperref}
\usepackage{url}
\usepackage{amsthm}
\usepackage{bbm}
\usepackage{physics}
\usepackage{algorithm}
\usepackage{algpseudocode}
\usepackage{caption}
\usepackage{microtype}
\usepackage{graphicx}
\usepackage{wrapfig}
\usepackage{subfigure}
\usepackage{booktabs}
\newcommand\Algphase[1]{%
\vspace*{-.7\baselineskip}\Statex\hspace*{\dimexpr-\algorithmicindent-2pt\relax}\rule{\columnwidth}{0.4pt}%
\Statex\hspace*{-\algorithmicindent}\textbf{#1}%
\vspace*{-.7\baselineskip}\Statex\hspace*{\dimexpr-\algorithmicindent-2pt\relax}\rule{\columnwidth}{0.4pt}%
}

\newtheorem{definition}{Definition}
\newtheorem{theorem}{Theorem}
\newtheorem{proposition}{Proposition}

\title{Wasserstein-regularized conformal \\prediction under general distribution shift}

\author{
Rui Xu,  Sihong Xie\thanks{Corresponding author:
Sihong Xie, \texttt{sihongxie@hkust-gz.edu.cn}}\\
The Hong Kong University of Science and Technology (Guangzhou)\\
\texttt{rxu233@connect.hkust-gz.edu.cn, sihongxie@hkust-gz.edu.cn} \\
\And
Chao Chen\\
Harbin Institute of Technology\\
\texttt{cha01nbox@gmail.com}
\And
Yue Sun, Parvathinathan Venkitasubramaniam\\
Lehigh University\\
\texttt{yus516@lehigh.edu, pav309@lehigh.edu}
}
\iclrfinalcopy 
\begin{document}

\maketitle
\begin{abstract}
Conformal prediction yields a prediction set with guaranteed $1-\alpha$ coverage of the true target under the i.i.d. assumption, 
which may not hold and lead to a gap between $1-\alpha$ and the actual coverage. 
Prior studies bound the gap using total variation distance, 
which cannot identify the gap changes under distribution shift at a given $\alpha$. 
Besides, existing methods are mostly limited to covariate shift,
while general joint distribution shifts are more common in practice but less researched.
In response, we first propose a Wasserstein distance-based upper bound of the coverage gap and analyze the bound using probability measure pushforwards between the shifted joint data and conformal score distributions, enabling a separation of the effect of covariate and concept shifts over the coverage gap. We exploit the separation to design an algorithm based on importance weighting and regularized representation learning (WR-CP) to reduce the Wasserstein bound with a finite-sample error bound.
WR-CP achieves a controllable balance between conformal prediction accuracy and efficiency.
Experiments on six datasets prove that WR-CP can reduce coverage gaps to $3.2\%$ across different confidence levels and outputs prediction sets 37$\%$ smaller than the worst-case approach on average.

\end{abstract}

\section{Introduction}
Because of data noise, unobservable factors, and knowledge gaps, stakeholders must also consider prediction uncertainty in machine learning applications, especially in areas such as fintech~\citep{ryu2020sustainable}, healthcare~\citep{feng2021review}, and autonomous driving~\citep{seoni2023application}. Conformal prediction (CP) addresses prediction uncertainty by generating a set of possible targets instead of a single prediction~\citep{vovk2005algorithmic, shafer2007tutorialconformalprediction, angelopoulos2021gentle}. We focus on CP in \textbf{regression tasks}. With a trained model $h$, CP calculates the difference (\textit{conformal score}) between the predicted and actual target via a score function $s(x,y)=|h(x)-y|$ over some calibration instances. With the empirical $1-\alpha$ quantile $\tau$ of the conformal scores, the prediction set $C(x)$ of a test input $x$ contains all targets whose scores are smaller than $\tau$. If calibration and test data are independent and identically distributed (i.i.d.), the probability that the prediction set $C(x)$ contains the true target $y$ of $x$ is close to $1-\alpha$ (i.e. the \textit{coverage guarantee}).

\begin{figure*}[t]
\vspace{-15pt}
\centering
\captionsetup{singlelinecheck = false, justification=justified}
  \includegraphics[scale=0.3]{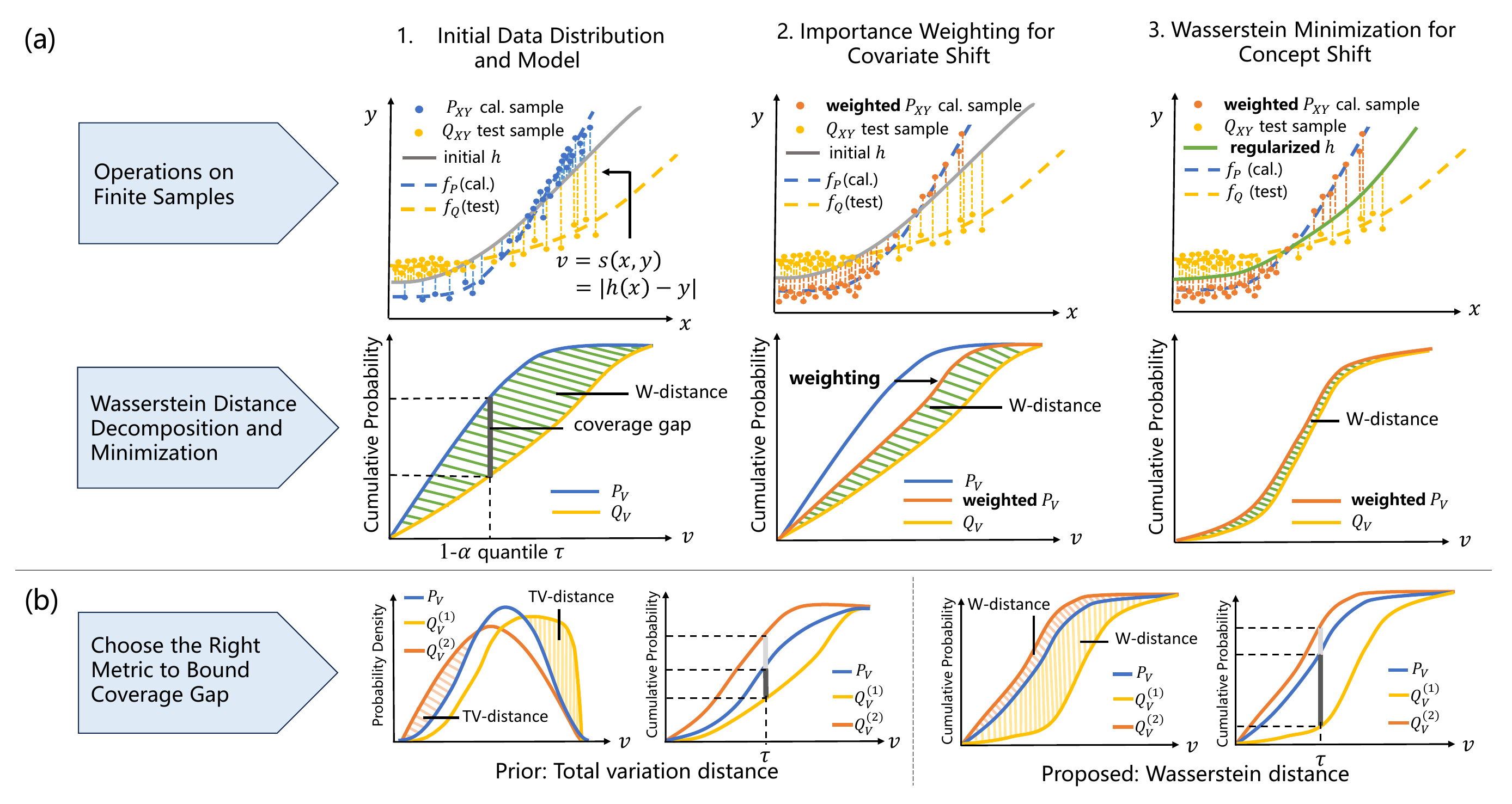}
  \caption{\footnotesize
  (a) Joint distribution shift can include both covariate shift ($P_X\neq Q_X$) and concept shift ($f_P\neq f_Q$). 
  Coverage gap (Eq.~(\ref{eq: coverage gap})) is the absolute difference in cumulative probabilities of calibration and test conformal scores at the $1-\alpha$ quantile $\tau$. We address covariate-shift-induced Wasserstein distance by applying importance weighting~\citep{tibshirani2019conformal} to calibration samples, and further minimize concept-shift-induced Wasserstein distance to obtain accurate and efficient prediction sets;
  (b) $Q_V^{(1)}$ and $Q_V^{(2)}$ are two distinct test conformal score distributions. Wasserstein distance (Eq.~(\ref{eq: Wasserstein distance})) integrates the vertical gap between two cumulative probability distributions overall \textit{all} quantiles, and is sensitive to coverage gap changes at \textit{any} quantile. Total variation distance fails to indicate coverage gap changes thoroughly as it is agnostic about where two distributions diverge.}
  \vspace{-20pt}
  \label{fig: proposed method} 
\end{figure*}

Denote $P_{XY}$ and $Q_{XY}$ the calibration and test distributions, respectively, in space $\mathcal{X}\times\mathcal{Y}$. We assume $y|x\sim N\left(f_P(x),\varepsilon_P\right)$ for $(x,y)\sim P_{XY}$ and $y|x\sim N\left(f_Q(x),\varepsilon_Q\right)$ for $(x,y)\sim Q_{XY}$.
In practice, the i.i.d. assumption can be violated by a joint distribution shift such that $P_{XY}\neq Q_{XY}$,
due to
a covariate shift ($P_X\neq Q_X$), a concept shift ($f_P\neq f_Q$), or both (Figure~\ref{fig: proposed method}(a) left)~\citep{kouw2018introduction}.  
With a distribution shift, the coverage guarantee fails, leading to a gap between the probability that $y\in C(x)$ and $1-\alpha$. 
Formally, denoting $P_V$ and $Q_V$ the calibration and test conformal score distributions, respectively, the coverage gap is the difference between the cumulative density functions (CDFs) of $P_V$ and $Q_V$ at quantile $\tau$ (Figure~\ref{fig: proposed method}(a) left).
Prior methods are concerned with the worst-case shifts and 
passively expand prediction sets as much as possible to meet the coverage guarantee for \textit{any} shifted test distribution, leading to excessively large and inefficient prediction sets ~\citep{gendler2021adversarially,cauchois2024robust,zou2024coverage,yan2024provably}.
Recent works assume knowledge about the distribution shifts between test and calibration distribution~\citep{barber2023conformal,angelopoulos2022conformal,colombo2024normalizing}. The knowledge is further embedded as the total variation (TV) distance between conformal score distributions $P_V$ and $Q_V$ to bound and minimize the coverage gap. 
However, the TV distance ignores where two conformal score distributions differ, while the coverage gap is defined at a specific $\alpha$ and is location-dependent, making TV distance less indicative of coverage gap during model optimization (Figure~\ref{fig: proposed method}(b) right).

Opposing to TV distance,
we adopt Wasserstein distance over the space of probability distributions of conformal score to upper bound the coverage gap under joint distribution shift.
Such an upper bound integrates the vertical gap between the CDFs of two conformal score distributions $P_V$ and $Q_V$ and measures the gap 
at \textit{any} $\alpha$ 
(Figure~\ref{fig: proposed method}(b) left),
indicating coverage gap at a \textit{given} $\alpha$
for distribution discrepancy minimization and coverage guarantee (Section~\ref{sec:upper-bounding-gap}, Appendix~\ref{appendix: TV v.s. W}).
Targeting more effective algorithms specifically for covariate and concept shifts that constitute joint distribution shift,
we further penetrate the complex landscape of joint distributional shift.
We disentangle the complex dependencies between the Wasserstein upper bound and covariate and concept shifts using a novel pushforwards of probability measure,
decomposing the bound into two Wasserstein terms so that the effects of covariate and concept shifts on the coverage gap are independent~(Eq.~(\ref{eq: Wasserstein triangle in decomposition})).
Theoretical analyses crystalize the link between CP coverage gap, the smoothness of the conformal residue and predictive model, and the amount of covariate and concept shifts (Section~\ref{sec:upper-bound-smoothness}).
The decomposition allows representation learning using 
importance weighting~\citep{tibshirani2019conformal} that reduces the covariate-shift-induced term,
and minimization of the concept-shift-induced term (Figure~\ref{fig: proposed method}(a)) with finite samples 
with an empirical error bound (Section~\ref{sec:empirical-bound}).
We proved the effectiveness of the resulting algorithm,
Wasserstein-regularized conformal prediction (WR-CP),
for multi-source domain generalization where the test distribution is an unknown mixture of training distributions.
On six datasets from applications including AI4S~\citep{misc_airfoil_self-noise_291}, smart transportation~\citep{cui2019traffic,guo2019attention}, and epidemic spread forecasting~\citep{deng2020cola},
experiments across various $\alpha$ values (0.1 to 0.9) demonstrate that coverage gaps are reduced to $3.2\%$ and the prediction set sizes are 37$\%$ smaller than those generated by the worst-case approach on average.
Besides, WR-CP allows a smooth balance between prediction coverage and efficiency (Figure~\ref{fig: Pareto front}).

\section{Background and related works}
\subsection{Conformal prediction}
Let $X\in\mathcal{X}\subseteq\mathbb{R}^d$ and $Y\in\mathcal{Y}\subseteq\mathbb{R}$ denote the input and output random variable, respectively.
A hypothesis $h: \mathcal{X}\rightarrow\mathcal{Y}$ is a model trained to predict target $Y$ from feature $X$. 
We observe $n$ instances $(X_1,Y_1),...,(X_n,Y_n)$ from calibration distribution $P_{XY}$. Taking $(x,y)$ as a realization of $(X,Y)$, a score function $s(x,y):\mathcal{X}\times\mathcal{Y}\rightarrow\mathcal{V}\subseteq\mathbb{R}$ quantifies how $(x,y)$ conforms to the model $h$. For regression tasks, typically $s(x,y)=|h(x)-y|$. Split conformal prediction is widely-used, and defines calibration conformal scores $V_i=s(X_i,Y_i)$ for $i=1,...,n$~\citep{papadopoulos2002inductive}. Letting $\tau\in\mathcal{V}$ be the ${\lceil(1-\alpha)(n+1)\rceil}/{n}$ quantile of $V_1,...,V_n$,
the prediction set of input $X_{n+1}$ is
\begin{equation}
\label{eq: prediction set}
    C(X_{n+1})=\left\{\hat{y}:s(X_{n+1},\hat{y})\leq\ \tau, \hat{y}\in\mathcal{Y}\right\}.
\end{equation}
Consider the instance $(X_{n+1}, Y_{n+1})$ following a test distribution $Q_{XY}$. If the test and calibration instances are i.i.d. (i.e. $P_{XY}=Q_{XY}$) , the probability that the true target $Y_{n+1}$ is included in $C(X_{n+1})$ is at least $1-\alpha$. If calibration conformal scores are almost surely distinct, we can also bound the probability from above by $1-\alpha+1/(n+1)$~\citep{angelopoulos2021gentle}. The bounded probability is called \textit{coverage guanrantee}:
\begin{equation}
\label{eq: coverage guarantee}
    \text{Pr}\left(Y_{n+1}\in C(X_{n+1})\right)\in \left[1-\alpha, 1-\alpha+{1}/({n+1})\right).
\end{equation}
\cite{vovk2005algorithmic} proved that the assumption of i.i.d instances can be relaxed to exchangeability of calibration and test instances. With exchangeability, prior CP methods proposed to improve the adaptiveness of prediction set to different test inputs~\citep{romano2019conformalized, romano2020classification, guan2023localized, amoukou2023adaptive,han2023splitlocalizedconformalprediction} and maintain conditional coverage guarantee for sub-populations of the test distribution~\citep{gibbs2023conformal, jung2022batch, feldman2021improving, cauchois2021knowing,foygel2021limits,stutz2021learning,einbinder2022training}. However, when the assumption is violated so that $P_{XY} \neq Q_{XY}$, coverage guarantee may not hold.

\subsection{Conformal prediction under distribution shifts}
\textbf{Covariate shift} ($P_X\neq Q_X$):~\cite{tibshirani2019conformal} adopted importance weighting by likelihood ratio between $P_X$ and $Q_X$ to satisfy the i.i.d assumption, so coverage is ensured under covariate shift. \textbf{Concept shift} ($f_P \neq f_Q$):~\cite{einbinder2022conformal,sesia2023adaptive} addressed CP under concept shift which is represented by label noise.

\textbf{Joint distribution shift} ($P_{XY} \neq Q_{XY}$) consists of covariate shift ($P_X \neq Q_X$) and/or concept shift ($f_P \neq f_Q$)~\citep{kouw2018introduction}. \cite{barber2023conformal},~\cite{angelopoulos2022conformal}, and ~\cite{angelopoulos2021gentle} upper-bound coverage gap via total variation distance, but TV distance cannot identify gap changes at a fixed $\alpha$.
To reduce the gap, ~\cite{gibbs2021adaptive},~\cite{xu2021conformal}, and~\cite{gibbs2024conformal} focus on CP under dynamic shift (test distribution changes over time). Meanwhile, some works concentrate on static shift (test distribution unchanged). These works can be categorized into two pipelines. The first pipeline modifies vanilla CP upon a residual-driven model for robust coverage~\citep{gendler2021adversarially,cauchois2024robust,zou2024coverage}. The second pipeline incorporates a conformal-based loss during training to obtain robust and efficient prediction sets~\citep{yan2024provably}.
However, these works treat a joint distribution shift as a whole and adopt a worst-case principle for prediction.

In this work, we explore CP under multi-source domain generalization, which focuses on developing a model that generalizes effectively to \textbf{unseen test distributions} by leveraging the data from multiple source distributions~\citep{sagawa2019distributionally, krueger2021out}.  A related, yet distinct area is federated CP, which aims to train a model across decentralized data sources to perform well on a known test distribution (typically a uniformly weighted mixture of source distributions) without requiring centralization to ensure privacy. Regarding federated CP, FCP~\citep{lu2023federated} and FedCP-QQ~\citep{humbert2023one} aim for a coverage guarantee when the test and calibration samples are exchangeable from the same mixture. When exchangeability does not hold, DP-FedCP~\citep{plassier2023conformal} addresses scenarios where test samples are drawn from a single source distribution, assuming that only label shifts ($P_Y\neq Q_Y$) occur among the source distributions. Besides, CP with missing outcomes is studied by~\cite{liu2024multi} where the samples from the test distribution are accessible. The proposed WR-CP does not consider privacy but works on a more generalized setup: the test samples are drawn from an \textbf{unknown random mixture} where both concept and covariate shifts can occur among the source domains.

\section{Method}\label{sec: method}

\subsection{Upper-bounding coverage gap by Wasserstein distance}
\label{sec:upper-bounding-gap}
As shown in Figure~\ref{fig: proposed method}(b), Wasserstein distance can effectively indicate changes in coverage gap across different values of $\alpha$. We formally upper-bound coverage gap via Wasserstein distance.
Let $V\in\mathcal{V}\subseteq\mathbb{R}$ be the random variable of conformal score. $P_V$ and $Q_V$ are calibration and test conformal score distributions, respectively. 
The guarantee in Eq.~(\ref{eq: coverage guarantee}) indicates that $\text{Pr}\left(s(X_{n+1},Y_{n+1})\leq \tau\right)\in [1-\alpha, 1-\alpha+1/(n+1))$. $F_{P_V}$ and $F_{Q_V}$ are CDFs of $P_V$ and $Q_V$, respectively. Under the i.i.d. assumption, $P_V=Q_V$, and thus $F_{Q_V}(\tau)=F_{P_V}(\tau) \in[ 1-\alpha,1-\alpha+1/(n+1))$. 
However, the assumption can be violated by a joint distribution shift, which may results in $P_{V}\neq Q_V$. In this case, $F_{P_V}(\tau)$ is still bounded, but $F_{Q_V}(\tau) \neq F_{P_V}(\tau)$. Inadequate coverage renders prediction sets unreliable, while excessive coverage leads to large prediction sets, reducing prediction efficiency, and we define coverage gap as the absolute difference\footnote{In this study, we assume that $n$ is sufficiently large for $F_{P_V}(\tau)$ to be approximated as $1-\alpha$, allowing us to view Eq.~(\ref{eq: coverage gap}) as the difference between $F_{Q_V}(\tau)$ and $1-\alpha$. }: 
\begin{equation}\label{eq: coverage gap}
    \text{Coverage gap}=|F_{P_V}(\tau)-F_{Q_V}(\tau)|.
\end{equation}

\begin{definition}[Kolmogorov Distance]\citep{gaunt2023bounding}
\label{def: Kolmogorov Distance}
$F_\mu$ and $F_\nu$ are the CDFs of probability measures $\mu$ and $\nu$ on $\mathbbm{R}$, respectively. Kolmogorov distance between $\mu$ and $\nu$ is given by $K(\mu,\nu)=\sup_{x\in\mathbbm{R}}|F_\mu(x)-F_\nu(x)|$.
\end{definition}
With Definition~\ref{def: Kolmogorov Distance}, as $\tau\in\mathcal{V}\subseteq\mathbbm{R}$, Eq.~(\ref{eq: coverage gap}) is bounded by $K(P_V,Q_V)$:
\begin{equation}\label{eq: bound by Kolmogorov distance}
    \text{Coverage gap}=|F_{P_V}(\tau)-F_{Q_V}(\tau)|\leq\sup\nolimits_{v\in\mathcal{V}}|F_{P_V}(v)-F_{Q_V}(v)|=K(P_V,Q_V).
\end{equation}
\begin{definition}[$p$-Wasserstein Distance]\citep{panaretos2019statistical}
    Given two probability measures $\mu$ and $\nu$ on a metric space $(\mathcal{X},c_\mathcal{X})$, where $\mathcal{X}$ is a set and $c_\mathcal{X}$ is a metric on $\mathcal{X}$, the Wasserstein distance of order $p\geq 1$ between $\mu$ and $\nu$ is 
    \begin{equation}\label{eq: Wasserstein distance}
        {W_p}(\mu,\nu)=\inf_{\gamma\in\Gamma(\mu,\nu)}\left(\int_{\mathcal{X}\times\mathcal{X}}{c_\mathcal{X}(x_1,x_2)^p\dd{\gamma(x_1,x_2)}}\right)^{1/p},
    \end{equation}
    where $\Gamma(\mu,\nu)$ is the set of all joint probability measures $\gamma$ on $\mathcal{X}\times\mathcal{X}$ with marginals $\gamma(\mathcal{A}\times\mathcal{X})=\mu(\mathcal{A})$ and $\gamma(\mathcal{X}\times \mathcal{B})=\nu(\mathcal{B})$ for all measurable sets $\mathcal{A},\mathcal{B}\subseteq\mathcal{X}$.
\end{definition}

\begin{proposition}~\citep{ross2011fundamentals}\label{pro: bound by Lebesgue density}
    If a probability measure $\mu$ in space $\mathbbm{R}$ has Lebesgue density bounded by $L$, then for any probability measure $\nu$,  $K(\mu,\nu)\leq\sqrt{2LW_1(\mu,\nu)}$.
\end{proposition}
In this work, let $W$ denote Wasserstein distance with $p=1$. Applying Eq.~(\ref{eq: bound by Kolmogorov distance}) and Proposition~\ref{pro: bound by Lebesgue density} with $L$ as the Lebesgue density bound of $P_V$, we can develop an upper bound by
\begin{equation}\label{eq: bound by Wasserstein distance}
    \text{Coverage gap}\leq\sqrt{2LW(P_V,Q_V)}.
\end{equation}
\subsection{Wasserstein distance decomposition and minimization}
In Eq.~(\ref{eq: bound by Wasserstein distance}), we show that the Wasserstein distance $W(P_V, Q_V)$ can effectively bound the coverage gap caused by a joint distribution shift. However, it is still not clear how the two components of joint distribution shift, namely, covariate shift in space $\mathcal{X}$ and concept shift in space $\mathcal{Y}$ lead to $W(P_V, Q_V)$ in space $\mathcal{V}$. Besides, we want the quantified contributions amenable to optimization techniques to reduce $W(P_V, Q_V)$. To the best of our knowledge, there is no prior work that suits this need. Therefore, we propose to upper-bound $W(P_V, Q_V)$ with two discrepancy terms due to covariate and concept shifts, and corresponding optimization methods to reduce $W(P_V, Q_V)$ via minimizing the two terms. 

\begin{definition}[Pushforward Measure]\label{def: pushforward}
    If $\mathcal{X}$ and $\mathcal{Y}$ are separate measurable spaces, $\mu$ is a prbability measure on $\mathcal{X}$, and $f:\mathcal{X}\rightarrow\mathcal{Y}$ is a measureable function, define the pushforward $f_\#\mu$ of $\mu$ through $f$ such that $f_\#\mu(\mathcal{A})=\mu(f^{-1}(\mathcal{A}))$ for all measurable set $\mathcal{A}\subseteq \mathcal{Y}$.
\end{definition}
With Definition~\ref{def: pushforward}, we have $P_Y=f_{P\#}P_{X}$ and $Q_Y=f_{Q\#}Q_{X}$. Besides, we define $s_P(x)=s(x,f_P(x))=|h(x)-f_P(x)|$ for $x\sim P_X$, and $s_Q(x)=s(x,f_Q(x))=|h(x)-f_Q(x)|$ for $x\sim Q_X$, leading to pushforwards of the conformal score $P_V=s_{P\#}P_{X}$ and $Q_V=s_{Q\#}Q_{X}$.

\begin{wrapfigure}[13]{r}{4.5cm}
\centering
\vspace{-8pt}
\captionsetup{singlelinecheck = false, skip=5pt, justification=justified}
  \includegraphics[width=0.3\textwidth]{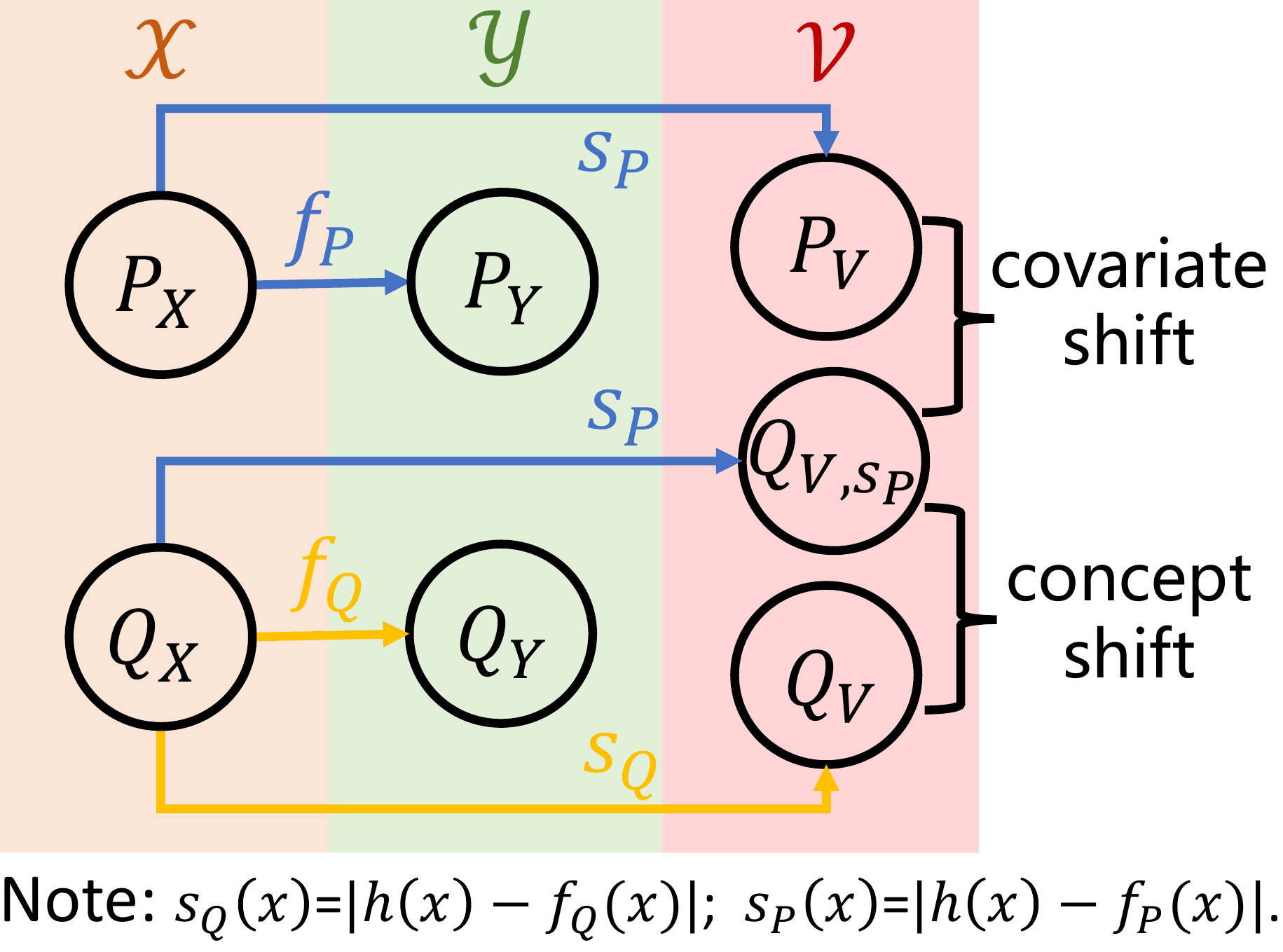}
  \caption{\footnotesize
  Pushforward measures.}
  \vspace{-10pt}
  \label{fig: pushforward} 
\end{wrapfigure}

To upper-bound $W(P_V, Q_V)$ about conformal scores according to covariate (concept, resp.) shifts in the $\mathcal{X}$ ($\mathcal{Y}$, resp.) space, we introduce a pushforward  $Q_{V,s_P}=s_{P\#}Q_{X}$ on $\mathcal{V}$. Since $P_V$ and $Q_{V,s_P}$ are pushforward measures by the same function $s_P$ from $P_X$ and $Q_X$, respectively,  $W(P_V,Q_{V,s_P})$ is a measure of covariate shift ($P_X\neq Q_X$). Also, as $Q_{V,s_P}$ and $Q_{V}$ are pushforward measures from the same source $Q_X$ by $s_P$ and $s_Q$, respectively, $W(Q_{V,s_P},Q_{V})$ can indicate the extent of concept shift ($f_P\neq f_Q$, and thus $s_P\neq s_Q$). The relationships among the pushforward measures are shown in Figure~\ref{fig: pushforward}. As~\cite{panaretos2019statistical} states, the triangle inequality holds that
\begin{equation}\label{eq: Wasserstein triangle in decomposition}
    W(P_V, Q_V)\leq W(P_V,Q_{V,s_P})+W(Q_{V,s_P},Q_{V}).
\end{equation}
With Eq.~(\ref{eq: Wasserstein triangle in decomposition}) bounding $W(P_V, Q_V)$, we design an approach to minimize the upper bound. First, we adopt importance weighting, which weights calibration conformal scores with the likelihood ratio ${\dd{Q_X(x)}}/{\dd{P_X(x)}}$.  \cite{tibshirani2019conformal} prove that importance weighting can preserve the coverage guarantee when only a covariate shift occurs. 
However, existing works do not include the weighting technique when dealing with a joint distribution shift.
We prove that importance weighting can minimize covariate-shift-induced Wasserstein distance, $W(P_V, Q_{V,s_P})$, even if a concept shift coincides.
Given any measurable set $\mathcal{A}\subseteq\mathcal{X}$, $\mathcal{B}:=\{s_P(x):x\in \mathcal{A}\}\subseteq\mathcal{V}$ . With Definition~\ref{def: pushforward},
\begin{equation*}
    \begin{split}
        &P_V(\mathcal{B})=\int_\mathcal{B}\dd{P_V(v)}=\int_\mathcal{B}\dd{(s_{P\#}P_X)(v)}=\int_\mathcal{A}\dd{P_X(x)}\quad\overrightarrow{\text{weighting}}\\&=\int_\mathcal{A}\frac{\dd{Q_X(x)}}{\dd{P_X(x)}}\dd{P_X(x)}=\int_\mathcal{A}\dd{Q_X(x)}=\int_\mathcal{B}\dd{(s_{P\#}Q_X)(v)}=\int_\mathcal{B}\dd{Q_{V,s_P}(v)}=Q_{V,s_P}(\mathcal{B}).
    \end{split}
\end{equation*}
Since importance weighting can transform $P_V$ to $Q_{V,s_P}$, $W(P_V, Q_{V,s_P})$ is minimized,
and the remaining term in the upper bound in Eq.~(\ref{eq: Wasserstein triangle in decomposition}) is the concept-shift-induced component $W(Q_{V,s_P}, Q_V)$. Next, we further minimize it during training, as illustrated in Figure~\ref{fig: proposed method}. The reasoning behind distinguishing between covariate and concept shifts is elaborated in Appendix~\ref{appendix sub: impact of weighting scores}.

\section{Theory}
\subsection{Upper-Bounding Wasserstein distance by covariate and concept shifts}
\label{sec:upper-bound-smoothness}
Although Eq.~(\ref{eq: Wasserstein triangle in decomposition}) upper bounds for $W(P_V,Q_V)$ by $W(P_V, Q_{V,s_P})$ and $W(Q_{V,s_P}, Q_V)$, it remains unclear how these shifts lead to these terms. Covariate shift can be more accurately quantified by $W(P_X, Q_X)$. Also, with $Q_{Y, f_P}=f_{P\#}Q_{X}$, $W(Q_{Y, f_P}, Q_Y)$ is a more direct way to measure concept shift by comparing $f_P$ and $f_Q$ based on $Q_X$. Therefore, we further upper-bound the two terms on the right-hand side of Eq.~(\ref{eq: Wasserstein triangle in decomposition}) using $W(P_X, Q_X)$ and $W(Q_{Y, f_P}, Q_Y)$.
We extend a theorem in \cite{aolaritei2022distributional}, which pushes two probability measures with the same function, while  Theorem~\ref{theorem: Wasserstein distance with pushforward} considers pushing with different functions.
\begin{theorem}\label{theorem: Wasserstein distance with pushforward} For probability measures $\mu$ and $\nu$ on metric space $(\mathcal{X},c_\mathcal{X})$, letting $f,g:\mathcal{X}\rightarrow\mathcal{Y}$ be measurable functions, $\mu_f$ and $\nu_g$ on metric space $(\mathcal{Y},c_\mathcal{Y})$ are pushforwards of $\mu$ and $\nu$ under functions $f$ and $g$, respectively. The Wasserstein distance between $\mu_f$ and $\nu_g$  holds the equivalence:
\begin{equation*}    W(\mu_f,\nu_g)=\inf_{\gamma'\in\Gamma(\mu_f,\nu_g)}\int\limits_{\mathcal{Y}\times\mathcal{Y}}c_\mathcal{Y}(y_1,y_2)\dd{\gamma'(y_1,y_2)}=\inf_{\gamma\in\Gamma(\mu,\nu)}\int\limits_{\mathcal{X}\times\mathcal{X}}c_\mathcal{Y}(f(x_1),g(x_2))\dd{\gamma(x_1,x_2)}.
\end{equation*}
\end{theorem}
As $\mathcal{V},\mathcal{Y}\subseteq\mathbb{R}$,  $c_\mathcal{V}(x_1,x_2)=c_\mathcal{Y}(x_1,x_2)=|x_1-x_2|$, Theorem~\ref{theorem: Wasserstein distance with pushforward} leads to the following
\begin{equation}
W(Q_{V,s_P},Q_{V})=\inf_{\gamma\in\Gamma(Q_X,Q_X)}\int_{\mathcal{X}\times\mathcal{X}}|s_P(x_1)-s_Q(x_2)|\dd{\gamma(x_1,x_2)},
\end{equation}
\begin{equation}
W(Q_{Y, f_P},Q_{Y})=\inf_{\gamma\in\Gamma(Q_X,Q_X)}\int_{\mathcal{X}\times\mathcal{X}}|f_P(x_1)-f_Q(x_2)|\dd{\gamma(x_1,x_2)}.
\end{equation}
Let $\gamma^*$ be the optimal transport plan of $W(Q_{Y, f_P},Q_{Y})$. With $\eta=\max\limits_{x_1,x_2\in\mathcal{X}}\frac{|s_P(x_1)-s_Q(x_2)|}{|f_P(x_1)-f_Q(x_2)|}$, we have
\begin{equation}\label{eq: concept shift bound}
\begin{split}
W(Q_{V,s_P},Q_{V})&\leq\int_{\mathcal{X}\times\mathcal{X}}|s_P(x_1)-s_Q(x_2)|\dd{\gamma^*(x_1,x_2)}\\&\leq\int_{\mathcal{X}\times\mathcal{X}}\eta|f_P(x_1)-f_Q(x_2)|\dd{\gamma^*(x_1,x_2)}=\eta W(Q_{Y, f_P},Q_{Y}).
\end{split}
\end{equation}
In Eq.~(\ref{eq: concept shift bound}), the first inequality holds as $\gamma^*$ may not be the optimal transport plan of $W(Q_{V,s_P}, Q_{V})$, and the second inequality follows the definition of $\eta$. Appendix~\ref{appendix: eta intuition} shows a geometric intuition of $\eta$.
\begin{theorem}\label{theorem: Wasserstein distance with Lipschitz continuity}
    For probability measures $\mu$ and $\nu$ on metric space $(\mathcal{X},c_\mathcal{X})$ with a measurable function $f:\mathcal{X}\rightarrow\mathcal{Y}$, $\mu_f$ and $\nu_f$ on metric space $(\mathcal{Y},c_\mathcal{Y})$ are the pushforward of $\mu$ and $\nu$ through function $f$, respectively. If $f$ has Lipschitz continuity constant $\kappa$, i.e., $\frac{c_\mathcal{Y}(f(x_1),f(x_2))}{c_\mathcal{X}(x_1,x_2)}\leq\kappa, \forall x_1,x_2\in\mathcal{X}$, 
    \begin{equation}
        W(\mu_f,\nu_f)\leq \kappa W(\mu,\nu).
    \end{equation}
\end{theorem}

As $\mathcal{X}\subseteq\mathbb{R}^d$, $c_\mathcal{X}(x_1,x_2)=\lVert x_1-x_2\rVert_2$. $\kappa$ is the Lipschitz constant of $s_P:\mathcal{X}\rightarrow\mathcal{V}$ such that $\frac{|s_P(x_1)-s_P(x_2)|}{\lVert x_1-x_2\rVert_2}\leq\kappa, \forall x_1,x_2\in\mathcal{X}$. With Theorem~\ref{theorem: Wasserstein distance with Lipschitz continuity}, as $P_V$ and 
$Q_{V,s_P}$ are pushforwards of $P_X$ and $Q_X$ through $s_P$, we have
\begin{equation}\label{eq: covariate shift bound}
    W(P_V,Q_{V,s_P})\leq\kappa W(P_X,Q_X).
\end{equation}

Plugging Eq.~(\ref{eq: concept shift bound}) and Eq.~(\ref{eq: covariate shift bound}) into Eq.~(\ref{eq: Wasserstein triangle in decomposition}), $W(P_V,Q_V)\leq\kappa W(P_X,Q_X)+\eta W(Q_{Y, f_P},Q_{Y})$. Therefore, by utilizing Eq.~(\ref{eq: bound by Wasserstein distance}), we can further bound the coverage gap using the magnitudes of covariate and concept shifts:
\begin{equation} \label{eq: bound gap by W-distance on X and Y}
    \text{Coverage gap}\leq\sqrt{2L\left(\kappa W(P_X,Q_X)+\eta W(Q_{Y, f_P},Q_{Y})\right)}.
\end{equation}
Equation (\ref{eq: bound gap by W-distance on X and Y}) highlights how covariate and concept shifts impact the coverage gap. While the values of $W(P_X,Q_X)$ and $W(Q_{Y, f_P},Q_{Y})$ are inherent properties of given data and cannot be altered, the parameters $\kappa$ and $\eta$ are linked to the model $h$, allowing minimizing $\kappa$ and $\eta$ via optimizing $h$. 
\subsection{Empirical upper bound of coverage gap}
\label{sec:empirical-bound}
In practice, $P_{V}$ and $Q_{V}$ are rarely available. Sometimes we may have access to their empirical distributions via the score function $s$, where $\hat{P}_{V}$ is derived from $n$ calibration samples and $\hat{Q}_{V}$ is obtained from $m$ test samples. Having the Wasserstein distance between the two empirical distributions $W(\hat{P}_V,\hat{Q}_V)$, we derive the error bound between the empirical form and $W(P_V,Q_V)$ by asymptotic estimation.

\begin{definition}[Upper Wasserstein Dimension]\citep{dudley1969speed}
    Given a set $\mathcal{A}\subseteq\mathcal{X}$, the $\epsilon$-covering number, denoted $\mathcal{N}_\epsilon(\mathcal{A})$, is the minimum $b$ such that $b$ closed balls, $\mathcal{B}_1,...,\mathcal{B}_b$, of diameter $\epsilon$ achieve $\mathcal{A}\subseteq\cup_{1\leq i\leq b}\mathcal{B}_i$. For a distribution $\mu$ in $\mathcal{X}$, the $(\epsilon,\tau)$-dimension is $d_\epsilon(\mu,\zeta)={-\log(\inf\{\mathcal{N}_\epsilon(\mathcal{A}):\mu(\mathcal{A})\geq1-\zeta\})}/{\log\epsilon}$. The upper Wassersteion dimension with $p=1$ is 
    \begin{equation}
        d_W(\mu)=\inf\{\varphi\in(2,\infty):\limsup\nolimits_{\epsilon\rightarrow0}d_\epsilon(\mu,\epsilon^{\frac{\varphi}{\varphi-2}})\leq \varphi\}.
    \end{equation}
\end{definition}
With the definition of upper Wasserstein dimension, \cite{weed2019sharp} conducted how an empirical distribution converges to its population by the Wasserstein distance between them.
\begin{proposition}\citep{weed2019sharp} \label{pro: Wasserstein distance converge}
    Given a probability measure $\mu$, $\sigma>d_W(\mu)$. If $\hat{\mu}_n$ is an empirical measure corresponding to $n$ i.i.d. samples from $\mu$, $\exists\lambda\in\mathbb{R}$ such that $\mathbb{E}[W(\mu,\hat{\mu}_n)]\leq\lambda n^{-1/\sigma}$. Furthermore, for $t>0$, $\textup{Pr}(W(\mu,\hat{\mu}_n)\geq\mathbb{E}[W(\mu,\hat{\mu}_n)]+t)\leq e^{-2nt^2}$. 
\end{proposition}
\begin{theorem}\label{theorem: Wasserstein approximation}
    Given two probability measures $\mu$ and $\nu$, $\sigma_\mu>d_W(\mu)$ and $\sigma_\nu>d_W(\nu)$. $\hat{\mu}_n$ and $\hat{\nu}_m$ are empirical measures corresponding to $n$ i.i.d. samples from $\mu$ and $m$ i.i.d. samples from $\nu$, respectively. For $t_\mu,t_\nu>0$, $\exists\lambda_\mu, \lambda_\nu\in\mathbb{R}$ with probability at least $(1-e^{-2n{t_\mu}^2})(1-e^{-2m{t_\nu}^2})$ that 
    \begin{equation}\label{eq: empirical upper bound theorem}
        W(\mu,\nu)\leq W(\hat{\mu}_n,\hat{\nu}_m)+\lambda_\mu n^{-1/\sigma_\mu}+\lambda_\nu m^{-1/\sigma_\nu}+t_\mu+t_\nu.
    \end{equation}
\end{theorem}
Applying Theorem~\ref{theorem: Wasserstein approximation} to Eq.~(\ref{eq: bound by Wasserstein distance}), we derive an empirical upper bound of coverage gap. Specifically, if $P_V$ has Lebesgue density bounded by $L$, for $t_P, t_Q>0$, $\sigma_P>d_W(P_V)$, and $\sigma_Q>d_W(Q_V)$, $\exists\lambda_P, \lambda_Q\in\mathbb{R}$ with probability at least $(1-e^{-2n{t_P}^2})(1-e^{-2m{t_Q}^2})$ that
\begin{equation}\label{eq: empirical bound}
    \text{Coverage gap}\leq\sqrt{2L\left(W(\hat{P}_V,\hat{Q}_V)+\lambda_P n^{-1/\sigma_P}+\lambda_Q m^{-1/\sigma_Q}+t_P+t_Q\right)}.
\end{equation}


\section{Application to multi-source conformal prediction}~\label{sec: algorithm}
In this work, we consider the test distribution to be an unknown random mixture of multiple training distributions, referred to as multi-source domain generalization~\citep{sagawa2019distributionally}. As highlighted by~\cite{cauchois2024robust}, achieving 1-$\alpha$ coverage for each of the training distributions ensures that the coverage on test data remains at 1-$\alpha$ if the test distribution is any mixture of the training distributions. We apply the methodology outlined in Section~\ref{sec: method} to this scenario, namely multi-source conformal prediction. Given training distributions $D_{XY}^{(i)}$ for $i=1,..,k$, we require $Q_{XY}$ follows
\begin{equation}\label{eq: mixture}
    Q_{XY}\in\left\{\sum\nolimits_{i=1}^kw_iD_{XY}^{(i)}: w_1,...,w_k\geq 0,\sum\nolimits_{i=1}^kw_i=1\right\}.
\end{equation}
In other words, $Q_{XY}$ is an unknown random mixture of  $D_{XY}^{(i)}$ for $i=1,..,k$. Next, we introduce a surrogate of $W(Q_{V,s_P}, Q_V)$, allowing the minimization of $W(Q_{V,s_P}, Q_V)$ even when the test distribution $Q_{XY}$ is unknown in practice. With the score function $s(x,y)$ and $D_V^{(i)}=s_\#D_{XY}^{(i)}$,
\begin{equation}\label{eq: conformal score convexity}
    Q_V=s_\#Q_{XY}=s_\#\sum\nolimits_{i=1}^kw_iD_{XY}^{(i)}=\sum\nolimits_{i=1}^kw_is_\#D_{XY}^{(i)}=\sum\nolimits_{i=1}^kw_iD_{V}^{(i)}.
\end{equation}
By marginalizing out $Y$ in Eq.~(\ref{eq: mixture}), we obtain $Q_X=\sum_{i=1}^kw_iD_{X}^{(i)}$. Similar to Eq.~(\ref{eq: conformal score convexity}), with score function $s_P(x)$ and $D_{V,s_P}^{(i)}=s_{P\#}D_X^{(i)}$, $Q_{V,s_P}=s_{P\#}Q_X=\sum_{i=1}^kw_iD_{V,s_P}^{(i)}$.
\begin{theorem}\label{theorem: Wasserstein inequality by convexity}
    In space $\mathcal{X}\subseteq\mathbb{R}$, $\nu$ is a mixture distribution of multiple distributions $\nu^{(i)}$, $i=1,...,k$, such that $\nu=\sum_{i=1}^kw_i\nu^{(i)}$ with $w_1,...,w_k\geq 0,\sum_{i=1}^kw_i=1$. For any distribution $\mu$ on $\mathcal{X}$, Wasserstein distance has the inequality that $W(\mu,\nu)\leq \sum\nolimits_{i=1}^kw_i W(\mu,\nu^{(i)})$.
\end{theorem}
By Theorem~\ref{theorem: Wasserstein inequality by convexity}, $    W(Q_{V,s_P},Q_{V})\leq\sum\nolimits_{i=1}^kw_iW(Q_{V,s_P},D_{V}^{(i)})\leq\sum\nolimits_{i=1}^kw_i\sum\nolimits_{i=1}^kw_iW(D_{V,s_P}^{(i)},D_{V}^{(i)})$.
The inequality offers a surrogate of $W(Q_{V,s_P}, Q_{V})$. Even if $Q_{XY}$ is unknown, with uniformly distributed weights, we minimize the expectation of the surrogate with $w_i=1/k$ for $i=1,...,k$:
$\min \frac{1}{k}\sum\nolimits_{i=1}^kW(D_{V,s_P}^{(i)},D_{V}^{(i)})$. Besides reducing the coverage gap, we also want smaller prediction errors, so we include empirical risk minimization (ERM)~\citep{vapnik1991principles} during training. Hence, with a loss function $l$ and a parameterized model $h_\theta$, we merge the constant $1/k$ with a hyperparameter $\beta$, and introduce the objective function
\begin{equation}\label{eq: objective function in population}
\min_\theta\sum\nolimits_{i=1}^k\mathbb{E}_{(x,y)\sim D^{(i)}_{XY}}\left[l(h_\theta(x),y)\right]+\beta\sum\nolimits_{i=1}^k W(D_{V,s_P}^{(i)},D_{V}^{(i)}).
\end{equation}
We design Wasserstein-regularized Conformal Prediction (WR-CP) to optimize $h_\theta$ by Eq.~(\ref{eq: objective function in population}) with finite samples and generate prediction sets with small coverage gaps. $\mathcal{S}^{(i)}_{XY}$ is the sample set drawn from $D_{XY}^{(i)}$ for $i=1,...,k$, and $\mathcal{S}^{P}_{XY}$ is the sample set drawn from ${P}_{XY}$.  $\mathcal{S}^{Q}_{XY}$ is a test set containing samples from an unknown distribution $Q_{XY}$.
Algorithm~\ref{algorithm: WR-CP} shows the implementation of WR-CP. Kernel density estimation (KDE) is applied to obtain $\hat{P}_X$, $\hat{D}_X^{(i)}$, and $\hat{Q}_X$ for the calculation of likelihood ratios, whereas $\hat{D}_V^{(i)}$ and $\hat{D}_{V,s_P}^{(i)}$ are estimated as discontinuous, point-wise distributions to ensure differentiability during training. We show the details of distribution estimation in Appendix~\ref{appendix: Distribution Estimation}. As Algorithm~\ref{algorithm: WR-CP} indicates, in the prediction phase, WR-CP follows the inference procedure of importance-weighted conformal prediction (IW-CP) proposed by~\cite{tibshirani2019conformal}. When $\beta=0$, Eq.~(\ref{eq: objective function in population}) returns to empirical risk minimization, and thus WR-CP becomes  IW-CP.
\begin{algorithm}
    \footnotesize
\caption{Wasserstein-regularized Conformal Prediction (WR-CP)}\label{algorithm: WR-CP}
\begin{algorithmic}[1]
\Require {training set $\mathcal{S}^{(i)}_{XY}$ from distribution $D^{(i)}_{XY}$ for $i=1,...,k$; calibration set $\mathcal{S}^{P}_{XY}$ from $P_{XY}$; $N$ training epochs; model $h_\theta$; score function $s(x,y)=|h_\theta(x)-y|$; loss function $l$; balancing hyperparameter $\beta$.}
\Algphase{Training Phase:}
\State Obtain $\hat{P}_X$ and $\hat{D}^{(i)}_X$ for $i=1,...,k$ by kernel density estimation;
\For {$j=1$ to $N$}
\State {$\mathcal{S}_V^{P}=\{s(x,y):(x,y)\in\mathcal{S}_{XY}^{P}\}$};
\For {$i=1$ to $k$}
\State{Obtain $\hat{D}_V^{(i)}$ from $\mathcal{S}_V^{(i)}:=\{s(x,y):(x,y)\in\mathcal{S}_{XY}^{(i)}\}$ by point-wise distribution estimation;}
\State{Weight all $v\in\mathcal{S}_V^{P}$ with normalized $\frac{\dd{\hat{D}}_X^{(i)}\left(x\right)}{\dd{\hat{P}}_X\left(x\right)}$, where $x$ is the feature that $(x, y) \in \mathcal{S}_{XY}^{P}, s(x, y) = v$}
\State{Obtain $\hat{D}_{V,s_P}^{(i)}$ from the weighted $\mathcal{S}_V^{P}$ by point-wise distribution estimation;}
\EndFor
\State{Optimize $h_\theta$ by $\min_\theta \sum\nolimits_{i=1}^k\mathbb{E}_{(x,y)\in \mathcal{S}^{(i)}_{XY}}\left[l(h_\theta(x),y)\right]+\beta\sum\nolimits_{i=1}^kW(\hat{D}_{V,s_P}^{(i)},\hat{D}_{V}^{(i)})$;}
\EndFor
\Algphase{Prediction Phase:}
\State{Obtain $\hat{Q}_X$ by kernel density estimation;}
\State{$\mathcal{S}_V^{P}=\{s(x,y):(x,y)\in\mathcal{S}_{XY}^{P}\}$};
\State{Weight all $v\in\mathcal{S}_V^{P}$ with normalized $\frac{\dd{\hat{Q}}_X\left(x\right)}{\dd{\hat{P}}_X\left(x\right)}$, where $x$ is the feature that $(x, y) \in \mathcal{S}_{XY}^{P}, s(x, y) = v$;}
\State{$\tau$ = $1-\alpha$ quantile of the weighted $\mathcal{S}_V^{P}$;}
\For {$(x,y)\in\mathcal{S}^Q_{XY}$}
\State {$C(x)=\{\hat{y}:s(x,\hat{y})\leq\tau,\hat{y}\in\mathcal{Y}\}$};
\EndFor
\end{algorithmic}
\end{algorithm}
\section{Experiments}\label{sec: experiment}
\subsection{Datasets and models}
Experiments were conducted on six datasets: (a) the airfoil self-noise dataset~\citep{misc_airfoil_self-noise_291}; (b) Seattle-loop~\citep{cui2019traffic}, PeMSD4, PeMSD8~\citep{guo2019attention} for traffic speed prediction; (c) Japan-Prefectures, and U.S.-States~\citep{deng2020cola} for epidemic spread forecasting. $k=3$ for the airfoil self-noise dataset, and $k=10$ for the other five datasets. We conducted 10 sampling trials for each dataset. Within each trails, we sampled $\mathcal{S}^{(i)}_{XY}$ from each subset $i$, for $i=1,...,k$. Given that calibration and training data are commonly assumed to follow the same distribution in CP, we sampled $\mathcal{S}^{P}_{XY}$ from the union of the $k$ subsets. Additionally, we generated $10k$ test sets for each dataset in every trial.  A multi-layer perceptron (MLP) with an architecture of (input dimension, 64, 64, 1) was utilized in all experimental setups to maintain comparison fairness. The detailed information about datasets and sampling procedure is shown in Appendix~\ref{appendix_sub: Dataset}. The code of our work is released on \texttt{https://github.com/rxu0112/WR-CP}.

\subsection{Correlation between Wasserstein distance and coverage gap}
We demonstrated Wasserstein distance can indicate coverage gap changes across $\alpha$ from 0.1 to 0.9 comprehensively, as illustrated in Figure~\ref{fig: proposed method}(b).
Specifically, for each dataset, $h_\theta$ was optimized by empirical risk minimization. Then, we applied vanilla conformal prediction to each test set and calculated the average value of coverage gaps for $\alpha$ values from 0.1 to 0.9. Meanwhile, we also computed the Wasserstein distances between the calibration and each test conformal score distributions. Our findings highlighted a strong positive monotonic relationship between Wasserstein distance and the average value, indicating its sensitivity to coverage gap changes across different $\alpha$.

\textbf{Baselines.} Three baseline distance measures were selected. First of all, total variation (TV) distance was chosen as~\cite{barber2023conformal} aimed to use it to bound coverage gap.  Besides, Kullback-Leibler (KL)-divergence and expectation differenec ($\Delta\mathbb{E}$) were selected as they are widely applied in domain adaptation researches~\citep{nguyen2021kl,magliacane2018domain}.

\textbf{Metric.} We applied Spearman's coefficient, $-1\leq r_s\leq1$ to quantify the monotonic relationship between distance measures and the average coverage gap. The absolute value of the coefficient represents the strength of the correlation. Its sign indicates if a correlation is positive or negative. A \textbf{higher} positive $r_s$ means a \textbf{stronger} positive monotonic relation. We show the detailed definition of Spearman's coefficient in Appendix~\ref{appendix sub: Coeff}.

\textbf{Result.} Table~\ref{Table: Spearman Coeff} presents Spearman's coefficients between distance measures and the average coverage gap across the six datasets, with the standard deviations shown in parentheses. The highest coefficient is bold and the second-highest coefficient is underlined.
The result shows that the Wasserstein distance consistently exhibits a high coefficient, suggesting that Wasserstein distance is an effective indicator of the average coverage gap, and establishing it as a suitable optimization metric for maintaining coverage guarantees across various $\alpha$ values.
\begin{table}[!h]
\centering
\footnotesize
\captionsetup{justification=centering}
\caption{Spearman's coefficients between distance measures and the average coverage gap}
\def\arraystretch{1.2}
\begin{tabular}{c|c|c|c|c|c|c}
\toprule\toprule
Dataset & Airfoil& PeMSD4 & PeMSD8 & Seattle & U.S. & Japan\\ \midrule
$W$ & \textbf{0.59} (0.24) & \underline{0.84} (0.03) & \textbf{0.90} (0.03) & \textbf{0.84} (0.05) & \textbf{0.77} (0.06) & \textbf{0.57} (0.05) \\ 
TV & 0.45 (0.16) & \textbf{0.88} (0.03) & \underline{0.86} (0.06) & \underline{0.75} (0.09) & 0.67 (0.10) & 0.37 (0.06) \\ 
KL & 0.40 (0.21) & 0.49 (0.17) & 0.51 (0.09) & 0.45 (0.17) & 0.60 (0.11) & \underline{0.53} (0.05) \\ 
$\Delta\mathbb{E}$ & \underline{0.55} (0.19) & 0.78 (0.05) & 0.85 (0.04) & 0.71 (0.06) & \underline{0.68} (0.08) & 0.37 (0.09) \\ \bottomrule
\end{tabular}
\label{Table: Spearman Coeff}
\end{table}
\subsection{Evaluation of WR-CP in Wasserstein distance minimization}
We proved that WR-CP, utilizing importance weighting, can effectively minimize the Wasserstein distances resulting from both concept shift and covariate shift.

\textbf{Baselines.} Besides WR-CP, we also conducted vanilla CP and IW-CP on all sampled datasets.

\textbf{Metric.} These approaches were compared based on the Wasserstein distance between calibration and test conformal scores. To place greater emphasis on the vertical coverage gap between conformal score CDFs, the distances were normalized to mitigate the impact of varying score scales across datasets, enabling more meaningful comparisons.

\begin{figure*}[!h]
\centering
\captionsetup{singlelinecheck = false, justification=justified}
  \includegraphics[scale=0.4]{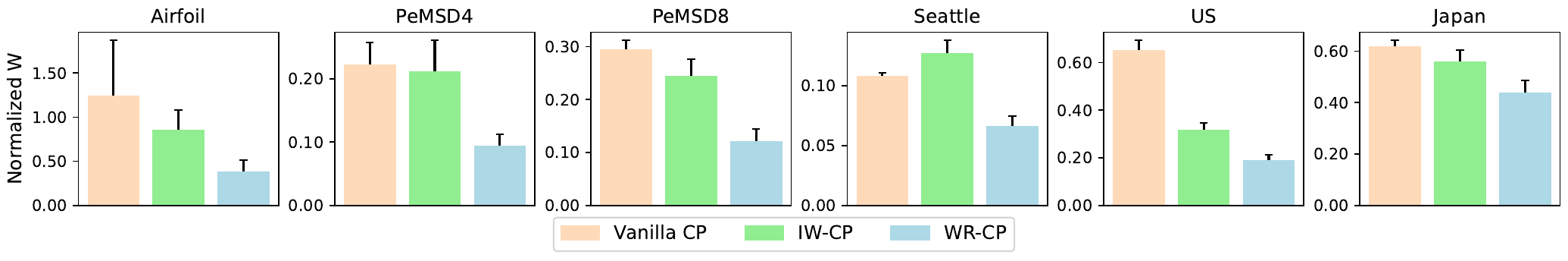}
  \caption{\footnotesize
  \textbf{Comparison of vanilla CP, IW-CP, and WR-CP based on normalized Wasserstein distance between calibration and test conformal scores:} IW-CP can only address the distance caused by covariate shift, while WR-CP reduces the distance from concept shift. The $\beta$ values for the WR-CP method are 9, 11, 9, 10, 13, and 13, respectively.}
  \vspace{0pt}
  \label{fig: comparison in Wasserstein distance} 
\end{figure*}

\textbf{Result.} Figure~\ref{fig: comparison in Wasserstein distance} shows that WR-CP consistently reduces Wasserstein distance. The extent of these reductions is dependent on the value of $\beta$. However, despite the ability to address covariate-shift-induced Wasserstein distance, importance weighting may not always lead to a reduction, as seen in the case of the Seattle-loop dataset. Further explanation of the phenomenon is provided in Appendix~\ref{appendix sub: impact of weighting scores}.

\subsection{Robust and efficient prediction sets by WR-CP}\label{subsec: coverage and size result}
We experimentally demonstrated that, compared with prior works, WR-CP is capable of reducing coverage gap without significantly sacrificing prediction efficiency. 

\textbf{Baselines.} Besides vanilla CP and IW-CP~\citep{tibshirani2019conformal}, conformalized quantile regression (CQR)~\citep{romano2019conformalized} was chosen as a representative method for adaptive CP.  We also included the worst-case conformal prediction (WC-CP), which is an implement of the worst-case approach proposed by~\cite{gendler2021adversarially,cauchois2024robust,zou2024coverage} in the convex hull setup.

\textbf{Metric.} We compared the coverage gaps and sizes of prediction sets generated by WR-CP and baselines as $\alpha$ ranges from 0.1 to 0.9 across all sampled datasets. Prediction sets are \textbf{better} when actual coverages are \textbf{more concentrated} around $1-\alpha$ and have \textbf{smaller} sizes.

\begin{figure*}[!t]
\centering
\captionsetup{singlelinecheck = false, justification=justified}
  \includegraphics[scale=0.4]{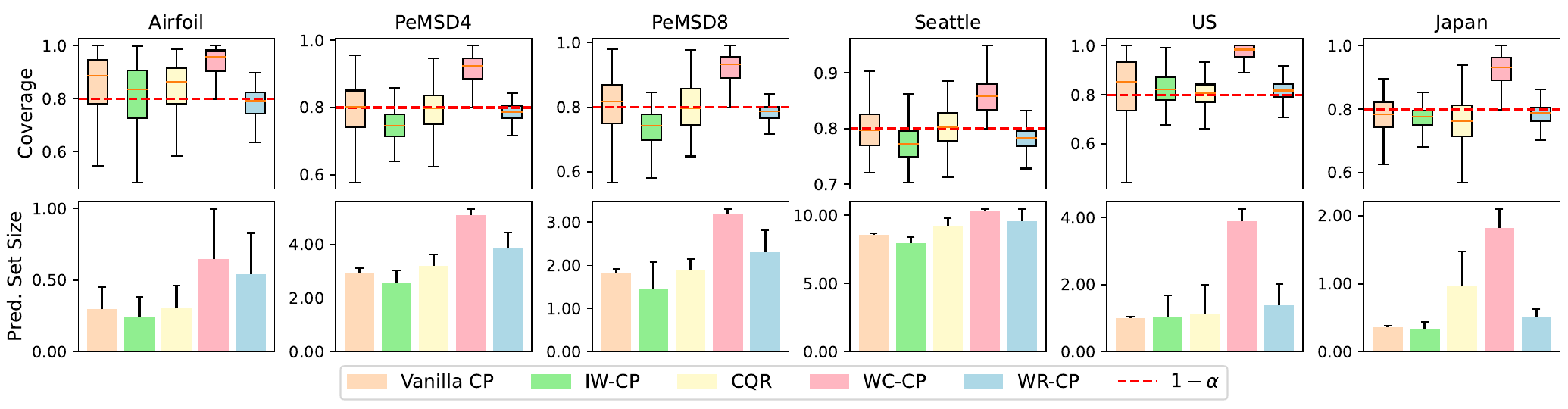}
  \caption{\footnotesize
  \textbf{Coverages and prediction set sizes of WR-CP and baselines with $\alpha=0.2$:} WR-CP makes coverages on test data more concentrated around the $1-\alpha$ level compared to vanilla CP, IW-CP, and CQR. While WC-CP ensures coverage guarantees, it leads to inefficient predictions due to large set sizes, whereas WR-CP mitigates this inefficiency. The $\beta$ values for the WR-CP method are 4.5, 9, 9, 6, 8, and 20, respectively.}
  \vspace{0pt}
  \label{fig: Coverage and Size 02} 
\end{figure*}

\textbf{Result.}
With $\alpha=0.2$, Figure~\ref{fig: Coverage and Size 02} confirms that WR-CP consistently exhibits the most concentrated coverages around $1-\alpha$ compared to vanilla CP, IW-CP, and CQR across datasets. While WC-CP maintains coverage guarantees under joint distribution shift, it leads to inefficient predictions. In contrast, WR-CP mitigates this inefficiency through smaller set sizes. We show the results with other $\alpha$ values in Appendix~\ref{appendix sub: additional result}. It is important to observe that vanilla CP and and IW-CP always have smaller prediction sets than WR-CP. Since WR-CP is trained with the additional Wasserstein regularization term in Eq.~(\ref{eq: objective function in population}), the trade-off inevitably causes an increase in prediction errors, which are proportional to conformal scores. Consequently, methods based on empirical risk minimization, like vanilla CP and IW-CP, tend to yield smaller prediction sets compared to WR-CP due to their lower conformal scores. We further discuss the trade-off in WR-CP in Subsection~\ref{subsec: ablation}. Lastly, we can see IW-CP have worse coverages than vanilla CP on Seattle-loop dataset, reflecting the fact that importance weighting enlarges Wasserstein distance on that dataset in Figure~\ref{fig: comparison in Wasserstein distance}. 
\subsection{Ablation study}\label{subsec: ablation}
\begin{figure*}[!t]
\centering
\captionsetup{singlelinecheck = false, justification=justified}
  \includegraphics[scale=0.4]{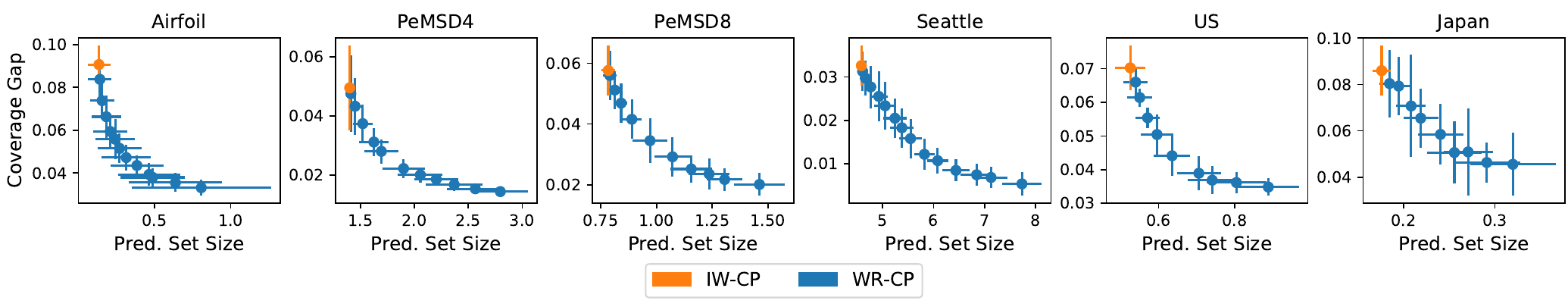}
  \caption{\footnotesize
  \textbf{Pareto fronts of coverage gap and prediction set size obtained from WR-CP with varying $\beta$:} WR-CP effectively balances conformal prediction accuracy and efficiency, providing a flexible and customizable solution. When $\beta=0$, WR-CP returns to IW-CP.}
  \vspace{0pt}
  \label{fig: Pareto front} 
\end{figure*}

As outlined in Eq.~(\ref{eq: objective function in population}), WR-CP is regulated by a hyperparameter $\beta$, which governs the trade-off between coverage gap and prediction set size. It is essential to investigate the performance of WR-CP under different $\beta$ values, which are listed in Appendix~\ref{appendix sub: beta values}. To achieve this, we conducted WR-CP on all sampled datasets with varying $\beta$ values. At each $\beta$ value, we calculated the average coverage gap and set size over $\alpha$ from 0.1 to 0.9. Finally, we obtained a Pareto front for each dataset in Figure~\ref{fig: Pareto front}. 
In particular, when $\beta=0$, WR-CP reverts back to IW-CP, so we emphasize the outcomes in this scenario as boundary solutions derived from IW-CP. The results indicate that WR-CP allows users to customize the approach based on their preferences for conformal prediction accuracy and efficiency.  We further explore whether WR-CP can achieve efficient prediction with a coverage guarantee in Appendix~\ref{appendix: prediction efficiency with coverage guarantee}. The limitations of our study are presented in Appendix~\ref{appendix: limitation}.
\section{Conclusion}
In this work, we point out that the coverage gap of conformal prediction under joint distribution shift relies on the distance between the CDFs of calibration and test conformal score distributions. Based on this observation, we propose an upper bound of coverage gap utilizing Wasserstein distance, offering better identifiability of gap changes at different $\alpha$. We conduct a detailed analysis of the bound by utilizing probability measure pushforwards from the shifted joint data distribution to conformal score distributions. This approach allows us to explore the separation of the impact of covariate and concept shifts on the coverage gap. Based on the separation, we design Wasserstein-regularized conformal prediction (WR-CP) via importance weighting and regularized representation learning, which can obtain accurate and efficient prediction sets with controllable balance. The performance of WR-CP is experimentally analyzed with diverse baselines and datasets. 
\newpage
\section*{Acknowledgment}
Sihong Xie was supported by the Department of Science and Technology of Guangdong Province (Grant No. 2023CX10X079), the National Key R\&D Program of China (Grant No. 2023YFF0725001), the Guangzhou-HKUST(GZ) Joint Funding Program (Grant No. 2023A03J0008), and Education Bureau Guangzhou Municipality.
\bibliography{iclr2025_conference}
\bibliographystyle{iclr2025_conference}

\newpage
\appendix
\section{Proofs of theorems}\label{appendix: Proofs}
\subsection{Proof of theorem~\ref{theorem: Wasserstein distance with pushforward}}
\begin{proof}
    We define $f\times g$ by $f\times g(x_1,x_2)=(f(x_1),g(x_2))=(y_1,y_2)$. Let $\text{Id}_\mathcal{X}$ be the identity mapping function on $\mathcal{X}$, and let $\pi_i$ be the mapping function to the $i$-th marginal. The proof follows Proposition 3 in the work by~\cite{aolaritei2022distributional}.
    
    First, we prove the inclusion that $(f\times g)\#\Gamma(\mu,\nu)\subset\Gamma(f_\#\mu,g_\#\nu)$. Consider $\gamma\in\Gamma(\mu,\nu)$, so it is equivalent to prove that $(f\times g)_\#\gamma\in\Gamma(f_\#\mu,g_\#\nu)$, which means the marginals of $(f\times g)_\#\gamma$ are $f_\#\mu$ and $g_\#\nu$. For any continuous and bounded function $\phi:\mathcal{Y}\rightarrow\mathbb{R}$, we have 
    \begin{equation}
        \begin{split}
            \int_{\mathcal{Y}\times\mathcal{Y}}{\phi(y_1)\dd{((f\times g)_\#\gamma)(y_1,y_2)}}=\int_{\mathcal{X}\times\mathcal{X}}{\phi(f(x_1))\dd{\gamma(x_1,x_2)}}\\
            =\int_\mathcal{X}{\phi(f(x_1))\dd{\mu(x_1)}}=\int_\mathcal{Y}{\phi(y_1)\dd{(f_\#\mu)(y_1)}},
        \end{split}
    \end{equation}
    so we obtain $\pi_{1\#}((f\times g)_\#\gamma)=f_\#\mu$ and similarly derive $\pi_{2\#}((f\times g)_\#\gamma)=g_\#\nu$.

    Secondly, we need to prove $\Gamma(f_\#\mu,g_\#\nu)\subset(f\times g)\#\Gamma(\mu,\nu)$. With $\gamma'\in\Gamma(f_\#\mu,g_\#\nu)$, we seek $\gamma\in\Gamma(\mu.\nu)$ such that $(f\times g)_\#\gamma=\gamma'$. To do so, let $\gamma_{12}:=(\text{Id}_\mathcal{X}\times f)_\#\mu\in\Gamma(\mu,f_\#\mu)$, $\gamma_{23}:=\gamma'\in\Gamma(f_\#\mu,g_\#\nu)$, and $\gamma_{34}:=(g\times \text{Id}_\mathcal{X})_\#\nu\in\Gamma(g_\#\nu,\nu)$. As $\pi_{2\#}\gamma_{12}=\pi_{1\#}\gamma_{23}=f_\#\mu$, and $\pi_{1\#}\gamma_{34}=\pi_{2\#}\gamma_{23}=g_\#\nu$, \cite{santambrogio2015optimal} ensures a joint probability measure $\Bar{\gamma}$ on $\mathcal{X}\times\mathcal{Y}\times\mathcal{Y}\times\mathcal{X}$ satisfying $(\pi_1\times\pi_2)_\#\Bar{\gamma}=\gamma_{12}$, $(\pi_2\times\pi_3)_\#\Bar{\gamma}=\gamma_{23}$, and $(\pi_3\times\pi_4)_\#\Bar{\gamma}=\gamma_{34}$. We demonstrate that $\gamma:=(\pi_1\times\pi_4)_\#\Bar{\gamma}$ is the probability measure we are seeking. For this, we prove $\gamma\in\Gamma(\mu,\nu)$ with any continuous and bounded function $\phi:\mathcal{X}\rightarrow\mathbb{R}$ by
    \begin{equation}\label{eq: prove gamma}
        \begin{split}
            \int_{\mathcal{X}\times\mathcal{X}}{\phi(x_i)\dd{\gamma(x_1,x_2)}}&=\int_{\mathcal{X}\times\mathcal{Y}\times\mathcal{Y}\times\mathcal{X}}\phi(x_1)\dd{\Bar{\gamma}(x_1,y_1,y_2,x_2)}\\&=\int_{\mathcal{X}\times\mathcal{Y}}\phi(x_1)\dd{\gamma_{12}(x_1,y_1)}=\int_{\mathcal{X}}\phi(x_1)\dd{\mu(x_1)}.
        \end{split}
    \end{equation}
    Eq.~(\ref{eq: prove gamma}) indicates $\pi_{1\#}\gamma=\mu$. Similarly, we can derive $\pi_{2\#}\gamma=\nu$. As a result, we can prove $(f\times g)_\#\gamma=\gamma'$ with any continuous and bounded function $\phi:\mathcal{Y}\times\mathcal{Y}\rightarrow\mathbb{R}$ by
    \begin{equation}
    \begin{split}
        &\int_{\mathcal{Y}\times\mathcal{Y}}{\phi(y_1,y_2)\dd{((f\times g)_\#\gamma)(x_1,x_2)}}\\&=\int_{\mathcal{X}\times\mathcal{X}}{\phi(f(x_1),g(x_2))\dd{\gamma(x_1,x_2)}}\\&=\int_{\mathcal{X}\times\mathcal{Y}\times\mathcal{Y}\times\mathcal{X}}{\phi(f(x_1),g(x_2))\dd{\Bar{\gamma}(x_1,y_1,y_2,x_2)}}\\&
        =\int_{\mathcal{X}\times\mathcal{Y}\times\mathcal{Y}\times\mathcal{X}}{\phi(y_1,y_2)\dd{\Bar{\gamma}(x_1,y_1,y_2,x_2)}}\\&=\int_{\mathcal{Y}\times\mathcal{Y}}{\phi(y_1,y_2)\dd{\gamma_{23}(y_1,y_2)}}=\int_{\mathcal{Y}\times\mathcal{Y}}{\phi(y_1,y_2)\dd{\gamma'(y_1,y_2)}}.
    \end{split}
    \end{equation}
    As $(f\times g)\#\Gamma(\mu,\nu)\subset\Gamma(f_\#\mu,g_\#\nu)$ and $\Gamma(f_\#\mu,g_\#\nu)\subset(f\times g)\#\Gamma(\mu,\nu)$, we obtain $(f\times g)\#\Gamma(\mu,\nu)=\Gamma(f_\#\mu,g_\#\nu)$.
    Finally, we prove Theorem~\ref{theorem: Wasserstein distance with pushforward} by
    \begin{equation}
        \begin{split}
            W(\mu_f,\nu_g)&=W(f_\#\mu,g_\#\nu)\\&=\inf_{\gamma'\in\Gamma(f_\#\mu,g_\#\nu)}c_\mathcal{Y}(y_1,y_2)\dd{\gamma'(y_1,y_2)}\\&=\inf_{\gamma'\in(f\times g)_\#\Gamma(\mu,\nu)}\int_{\mathcal{Y}\times\mathcal{Y}}c_\mathcal{Y}(y_1,y_2)\dd{\gamma'(y_1,y_2)}\\&=\inf_{\gamma\in\Gamma(\mu,\nu)}\int_{\mathcal{Y}\times\mathcal{Y}}c_\mathcal{Y}(y_1,y_2)\dd{((f\times g)_\#\gamma)(y_1,y_2)}\\&=\inf_{\gamma\in\Gamma(\mu,\nu)}\int_{\mathcal{Y}\times\mathcal{Y}}c_\mathcal{Y}(f(x_1),g(x_2))\dd{\gamma(y_1,y_2)}
        \end{split}
    \end{equation}
\end{proof}
\subsection{Proof of theorem~\ref{theorem: Wasserstein distance with Lipschitz continuity}}
\begin{proof}
    Let $\gamma'\in\Gamma(\mu_f,\nu_f)$ be the pushforward of $\gamma\in\Gamma(\mu,\nu)$ via function $f\times f$. We can apply Theorem~\ref{theorem: Wasserstein distance with pushforward} to $W(\mu_f,\nu_f)$ and obtain
    \begin{equation}
        W(\mu_f,\nu_f)=\inf_{\gamma\in\Gamma(\mu,\nu)}\int_{\mathcal{X}\times\mathcal{X}}c_\mathcal{Y}(f(x_1),f(x_2))\dd{\gamma(x_1,x_2)}.
    \end{equation}
    If the optimal transport plan for $W(\mu,\nu)$ is $\gamma^*$, and $\kappa$ bounds the Lipschitz continuity of $f$, we have 
    \begin{equation}\label{eq: Wasserstein distance with Lipschitz continuity}
    \begin{split}
                W(\mu_f,\nu_f)&\leq\int_{\mathcal{X}\times\mathcal{X}}c_\mathcal{Y}(f(x_1),f(x_2))\dd{\gamma^*(x_1,x_2)}\\&\leq\int_{\mathcal{X}\times\mathcal{X}}\kappa c_\mathcal{X}(x_1,x_2)\dd{\gamma^*(x_1,x_2)}=\kappa W(\mu,\nu).
    \end{split}
    \end{equation}
    In Eq.~(\ref{eq: Wasserstein distance with Lipschitz continuity}), the first inequality holds because $\gamma^*$ may not be the optimal transport plan for $W(\mu_f,\nu_f)$, and the second inequality holds due to the definition of $\kappa$.
\end{proof}

\subsection{Proof of theorem~\ref{theorem: Wasserstein approximation}}
\begin{proof}
As Wasserstein distance satisfies triangle inequality, $W(\mu,\nu)$ and $W(\hat{\mu}_n,\hat{\nu}_m)$ follow
\begin{equation}\label{eq: Wasserstein triangle ineqality}
     W(\mu,\nu)\leq W(\hat{\mu}_n,\mu) + W(\hat{\mu}_n,\nu) \leq W(\hat{\mu}_n,\mu) +  W(\hat{\mu}_n,\hat{\nu}_m) + W(\hat{\nu}_m,\nu).
\end{equation}
Given $\mathbb{E}[W(\mu,\hat{\mu}_n)]\leq\lambda_\mu n^{-1/\sigma_\mu}$ and $\mathbb{E}[W(\nu,\hat{\nu}_m)]\leq\lambda_\nu m^{-1/\sigma_\nu}$ from Proposition~\ref{pro: Wasserstein distance converge}, with probabilities at least $1-e^{-2n{t_\mu}^2}$ and $1-e^{-2m{t_\nu}^2}$, respectively, we have
\begin{equation}\label{eq: Wasserstein ineqality mu and nu}
    W(\mu,\hat{\mu}_n)\leq\lambda_\mu n^{-1/\sigma_\mu}+t_\mu, \text{  }W(\nu,\hat{\nu}_m)\leq\lambda_\nu m^{-1/\sigma_\nu}+t_\nu.
\end{equation}
It is reasonable to assume the two events in Eq.~(\ref{eq: Wasserstein ineqality mu and nu}) are independent, so we can apply them to Eq.~(\ref{eq: Wasserstein triangle ineqality}), and thus obtain Eq.~(\ref{eq: empirical upper bound theorem}) with probability at least $(1-e^{-2n{t_\mu}^2})(1-e^{-2m{t_\nu}^2})$.
\end{proof}
\subsection{Proof of theorem~\ref{theorem: Wasserstein inequality by convexity}}
\begin{proof}
    We denote $F_\mu$, $F_\nu$, and $F_{\nu^{(i)}}$ the corresponding CDFs of $\mu$, $\nu$, and $\nu^{(i)}$ for $i=1,...,k$.
    
    When two distributions are on the real number set $\mathbb{R}$ with Euclidean distance, $W$ of the two distributions equals the area between their CDFs. Therefore, the 1-Wasserstein distance between $\mu$ and $\nu$ is given by
        \begin{equation}\label{eq: Wasserstein on real number set}
            W(\mu,\nu)=\int_\mathcal{X}|F_\mu(x)-F_\nu(x)|\dd{x}.
        \end{equation}
    Since $\nu=\sum_{i=1}^kw_i\nu^{(i)}$, we have $F_\nu(x)=\sum_{i=1}^kw_iF_{\nu^{(i)}}(x)$. As $\nu$, $\nu^{(i)}$, and $\mu$ are definded on $\mathcal{X}\subseteq\mathbb{R}$, we can derive
    \begin{equation}
    \begin{split}
            W(\mu,\nu)&=\int_\mathcal{X}|F_\mu(x)-F_\nu(x)|\dd{x}
            =\int_\mathcal{X}\left|F_\mu(x)-\sum_{i=1}^kw_iF_{\nu^{(i)}}(x)\right|\dd{x}\\&=\int_\mathcal{X}\left|\sum_{i=1}^kw_iF_\mu(x)-\sum_{i=1}^kw_iF_{\nu^{(i)}}(x)\right|\dd{x}=\int_\mathcal{X}\left|\sum_{i=1}^kw_i\left(F_\mu(x)-F_{\nu^{(i)}}(x)\right)\right|\dd{x}\\&\leq\int_\mathcal{X}\sum_{i=1}^kw_i\left|F_\mu(x)-F_{\nu^{(i)}}(x)\right|\dd{x}=\sum_{i=1}^kw_i\int_\mathcal{X}\left|F_\mu(x)-F_{\nu^{(i)}}(x)\right|\dd{x}\\&=\sum_{i=1}^kw_iW(\mu,\nu^{(i)}).
    \end{split}
    \end{equation}
\end{proof}
\section{Comparison between total variation and wasserstein distance}
\label{appendix: TV v.s. W}
The total variation (TV) distance between two univariate distributions is defined as half of the absolute area between their probability density functions (PDFs). For instance, given two distributions $\mu$ and $\nu$ with PDFs $p_\mu$ and $p_\nu$, respectively, on space $\mathbb{R}_{\geq 0}$, the TV distance is given by
\begin{equation}\label{eq: def TVD}
    TV(\mu,\nu)=\frac{1}{2}\int_{\mathbb{R}_{\geq 0}}|p_\mu(x)-p_\nu(x)|\dd{x}.
\end{equation}
In contrast, we expand $W(\mu,\nu)$ according to Eq.~(\ref{eq: Wasserstein on real number set}) by
\begin{equation}\label{eq: Wasserstein expanded}
\begin{split}
        W(\mu,\nu)&=\int_{\mathbb{R}_{\geq 0}}|F_{\mu}(x)-F_{\nu}(x)|\dd{x}=\int_{\mathbb{R}_{\geq 0}}\left|\int_0^x p_\mu(t)\dd{t}-\int_0^x p_\nu(t)\dd{t}\right|\dd{x}\\& =\int_{\mathbb{R}_{\geq 0}}\left|\int_0^x p_\mu(t)- p_\nu(t)\dd{t}\right|\dd{x}.
\end{split}
\end{equation}
The inner integration between 0 and $x$ indicates Wasserstein distance cares where two distributions $\mu$ and $\nu$ differ, whereas the total variation distance in Eq.~(\ref{eq: def TVD}) does not take this into consideration. 

\begin{figure*}[h]
\centering
\captionsetup{singlelinecheck = false, justification=justified}
  \includegraphics[scale=0.35]{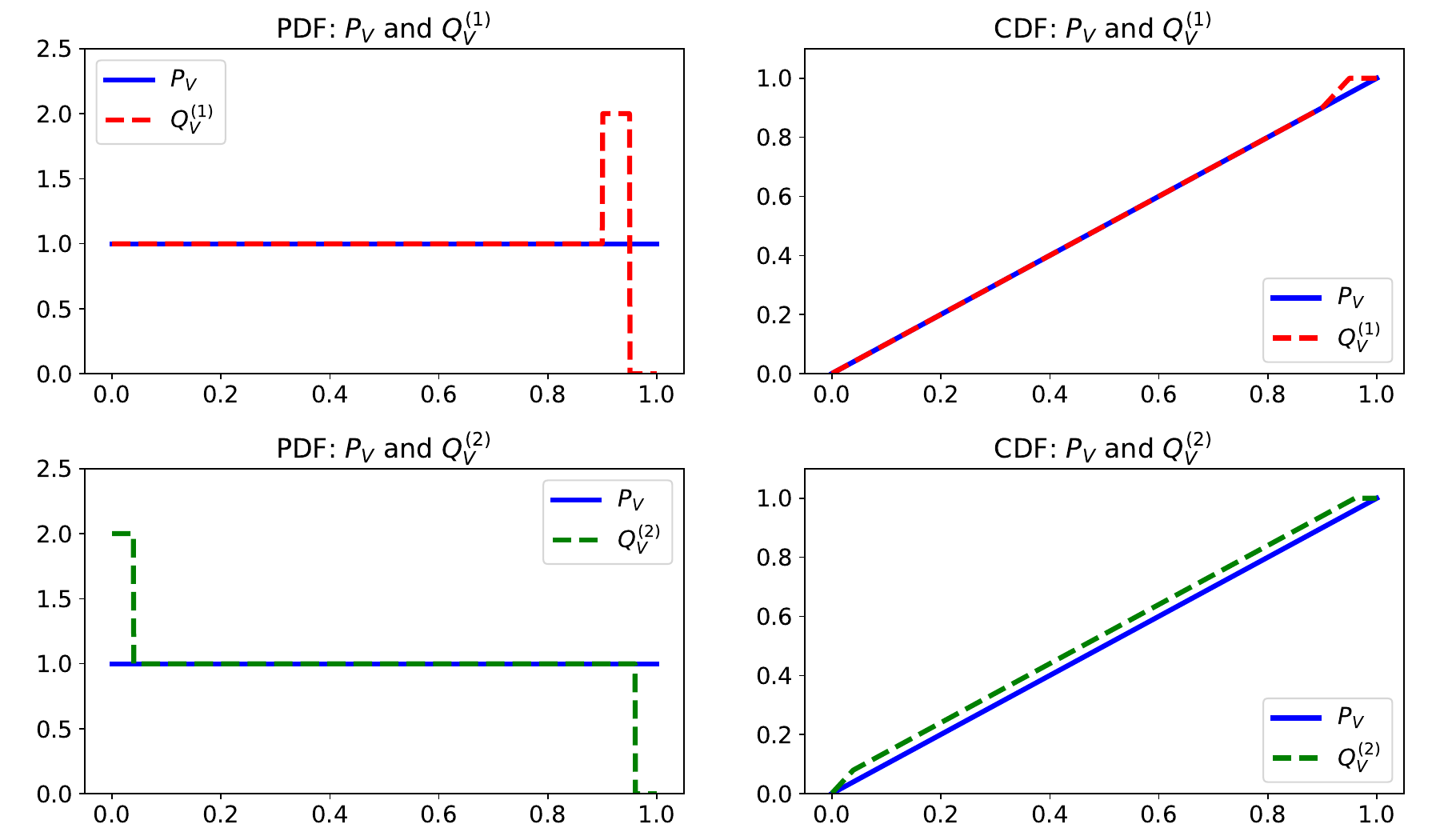}
  \caption{\footnotesize
  \textbf{Comparison between total variation distance and Wasserstein distance}: a reduction in the total variation distance does not necessarily result in CDFs becoming closer.}
  \vspace{0pt}
  \label{fig: Comparison bwt TVD and W}  
\end{figure*}

We would like to introduce a toy example to illustrate further why total variation distance can not consistently capture the closeness between two cumulative distribution functions (CDFs). Consider three conformal score distributions $P_V,Q_V^{(1)},Q_V^{(2)}$ on space $\mathbb{R}_{\geq 0}$ with their PDFs: 
\begin{equation*}
    p_{P_V^{ }}(v)=1,v\in[0,1];
\end{equation*}
\begin{equation*}
    p_{Q_V^{(1)}}(v)=
    \begin{cases}
    1 & \text{if $v\in[0,0.9]$},\\
    2 & \text{if $v\in(0.9,0.95]$};\\
    \end{cases}
\end{equation*}
\begin{equation*}
    p_{Q_V^{(2)}}(v)=
    \begin{cases}
    2 & \text{if $v\in[0,0.04]$},\\
    1 & \text{if $v\in(0.04,0.96]$}.\\
    \end{cases}
\end{equation*}
Therefore, we calculate $TV(P_V,Q_V^{(1)})=0.05$ and $TV(P_V,Q_V^{(2)})=0.04$, while $W(P_V,Q_V^{(1)})=0.0025$ and $W(P_V,Q_V^{(2)})=0.0384$. In this example, a reduction in total variation distance results in a larger Wasserstein distance between two CDFs. Intuitively, TVD only measures the overall difference between two distributions without accounting for the specific locations where they diverge. In contrast, the Wasserstein distance will be high when divergence occurs early (i.e., at a small quantile), especially if the discrepancy persists until the "lagging" CDF catches up. We visualize the example in Figure~\ref{fig: Comparison bwt TVD and W}.
\section{Rationale for differentiating covariate and concept shifts}\label{appendix sub: impact of weighting scores}
\begin{figure*}[!h]
\centering
\captionsetup{singlelinecheck = false, justification=justified}
  \includegraphics[scale=0.24]{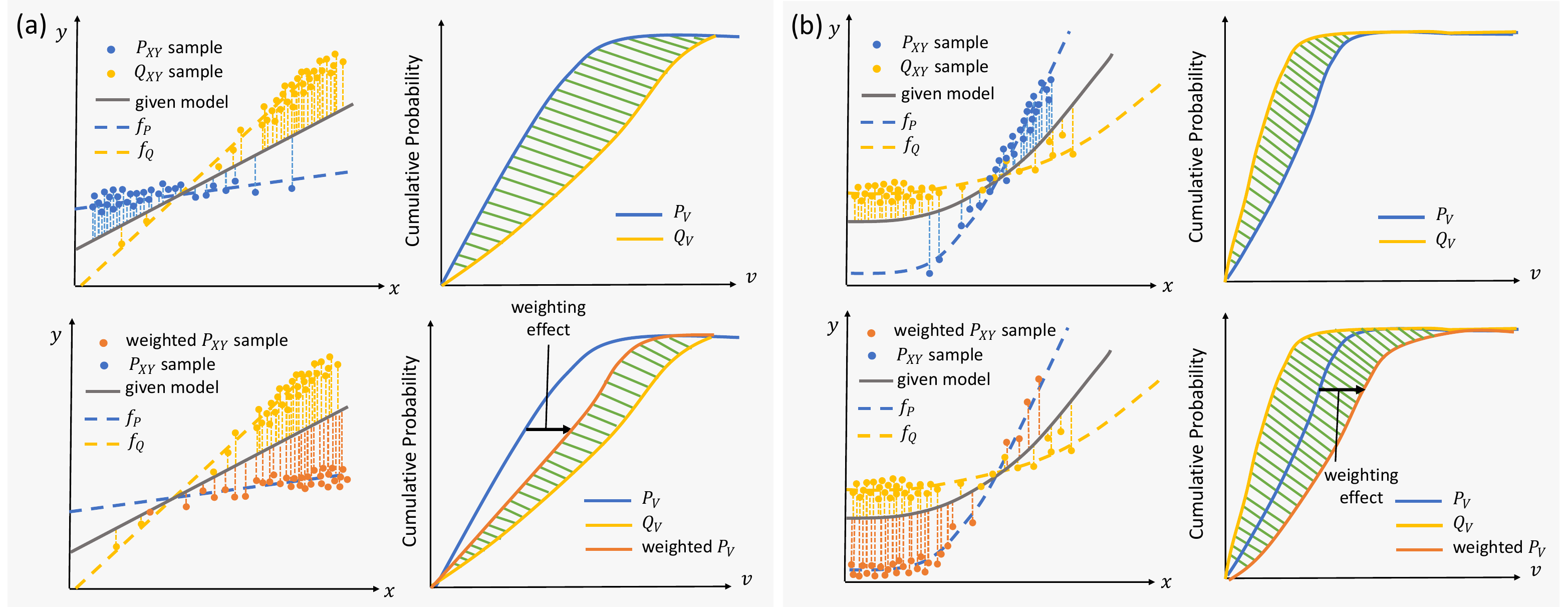}
  \caption{\footnotesize
  \textbf{Effect of importance weighting on Wasserstein distance:} (a) Scenario where importance weighting reduces Wasserstein distance; (b) Scenario where importance weighting enlarges Wasserstein distance.}
  \vspace{-10pt}
  \label{fig: weighting decrease and increase W} 
\end{figure*}

There are two key reasons to differentiate between covariate and concept shifts. First, making this distinction enables the application of importance weighting. Minimizing the Wasserstein regularization term inevitably increases prediction residuals. By applying importance weighting, we expect to reduce the distance, mitigating the adverse effects of regularization on optimizing the regression loss function in Eq.~(\ref{eq: objective function in population}). Figure~\ref{fig: comparison in Wasserstein distance} shows this expectation is met on five out of the six datasets. This occurs because, in most cases, covariate shifts exacerbate the distance caused by concept shifts ($f_P \neq f_Q$). Consequently, importance weighting effectively reduces this distance, as illustrated in Figure~\ref{fig: weighting decrease and increase W}(a) and evidenced by the results for the airfoil self-noise, PeMSD4, PeMSD8, U.S.-States, and Japan-Prefectures datasets in Figure~\ref{fig: comparison in Wasserstein distance}. However, there are instances where covariate shifts can alleviate the Wasserstein distance induced by concept shifts. In such cases, applying importance weighting may increase the distance, as demonstrated in the results for the Seattle-loop dataset in Figure~\ref{fig: comparison in Wasserstein distance}. This phenomenon is further illustrated in Figure~\ref{fig: weighting decrease and increase W}(b). 

Secondly, in multi-source CP, different training distributions $D _{XY}^{(i)}$ can suffer from different degrees of covariate and concept shifts. Importance weighting allows the regularized loss in Eq.~(\ref{eq: objective function in population}) to minimize the distance between training conformal score distribution $D _{V}^{(i)}$ and its correspondingly weighted calibration conformal score distribution $D _{V, s _P}^{(i)}$,  so the model can be more targeted on those whose remaining Wasserstein distances are large. Also, since various non-exchangeable test distributions will weight calibration conformal score distribution differently in the inference phase, prediction set sizes can be adaptive to different test distributions. In contrast, without importance weighting, the model can only regularize $\sum _{i=1}^k W(P _V, D _V^{(i)})$, and use the same quantile of $P_V$ to generate prediction sets for samples from all test distributions, resulting in the same prediction set size and lack of adaptiveness.

To further demonstrate the two reasons we mentioned above, we modify Wasserstein-regularization based on unweighted calibration conformal scores (i.e. $\sum _{i=1}^k W(P _V, D _V^{(i)})$) during training. Also, the weighting operation in the prediction phase in Algorithm \ref{algorithm: WR-CP} is removed accordingly. This method is denoted as WR-CP(uw). We performed WR-CP(uw) on the sampled data from the 10 trials of each dataset at $\alpha=0.2$ and compared its results with those of WR-CP. 

\begin{figure*}[!h]
\centering
\captionsetup{singlelinecheck = false, justification=justified}
  \includegraphics[scale=0.4]{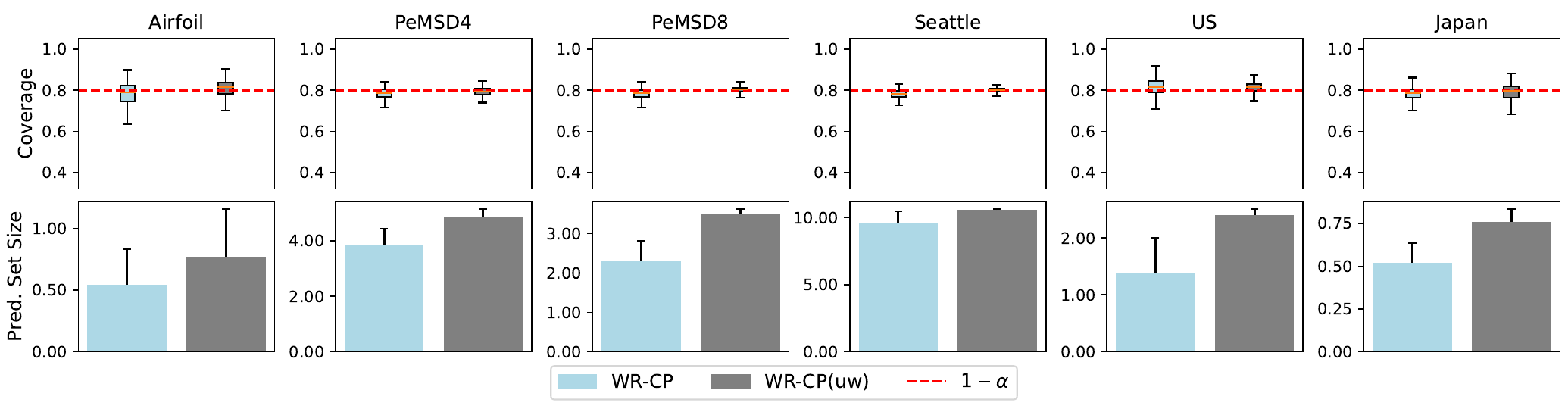}
  \caption{\footnotesize
  \textbf{Comparison between WR-CP and WR-CP(uw) at $\alpha=0.2$.} Both methods were implemented using the same $\beta$ values of 4.5, 9, 9, 6, 8, and 20 across the datasets.}
  \vspace{0pt}
  \label{fig: Comparison bwt WRCP and WRCP_uw 02} 
\end{figure*}

The comparison is depicted in Figure~\ref{fig: Comparison bwt WRCP and WRCP_uw 02}. Although the average coverage gaps between WR-CP and WR-CP(uw) are quite similar, at $3.1\%$ and $2.3\%$ respectively, the average prediction set size for WR-CP is $28.0\%$ smaller than that of WR-CP(uw). This observation proves our first reason that importance weighting effectively reduces the Wasserstein distance between calibration and test conformal scores. By doing so, it mitigates the side effect of optimizing the regularized objective function in Eq.~(\ref{eq: objective function in population}), which increases prediction residuals. Since larger residuals result in larger prediction sets, reducing residuals directly helps minimize prediction set size.
Additionally, the standard deviations of the prediction set sizes observed in WR-CP(uw) are typically smaller than those found in WR-CP. This proves the second reason that removing importance weighting will make prediction sets less adaptive to different test distributions.


\section{Geometric intuition of $\eta$}\label{appendix: eta intuition}
\begin{figure*}[h]
\centering
\captionsetup{singlelinecheck = false, justification=justified}
  \includegraphics[scale=0.35]{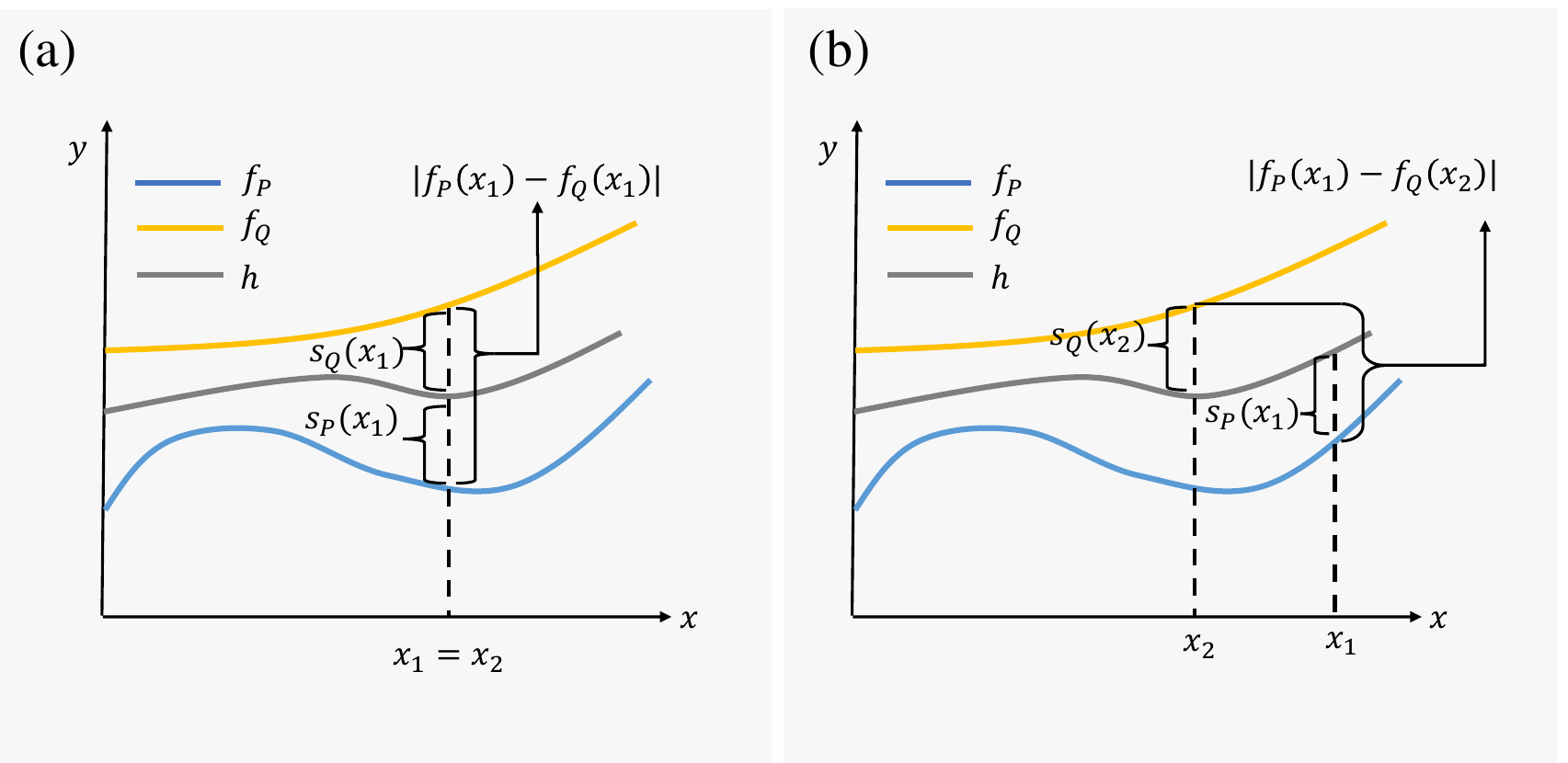}
  \caption{\footnotesize
  \textbf{Geometric intuition of $\eta$ when (a) $x_1=x_2$ and (b) $x_1\neq x_2$}: Intuitively, the residual difference caused by concept shift will be constrained by $\eta$.}
  \vspace{0pt}
  \label{fig: eta visuallization}  
\end{figure*}

To provide a geometric intuition of $\eta$, we expand the definition of $\eta$ as
\begin{equation}
\begin{split}
        \eta&= \max\limits_{x_1,x_2\in\mathcal{X}}\frac{|s_P(x_1)-s_Q(x_2)|}{|f_P(x_1)-f_Q(x_2)|}\\&=\max\limits_{x_1,x_2\in\mathcal{X}}\frac{|s\left(x_1,f_P(x_1)\right)-s\left(x_2,f_Q(x_2)\right)|}{|f_P(x_1)-f_Q(x_2)|}\\&=\max\limits_{x_1,x_2\in\mathcal{X}}\frac{\left||h(x_1)-f_P(x_1)|-|h(x_2)-f_Q(x_2)|\right|}{|f_P(x_1)-f_Q(x_2)|}.
\end{split}
\end{equation}
We first simplify the definition by assuming $x_1=x_2$, so the denominator is the absolute difference between two ground-truth mapping functions $f_P$ and $f_Q$ at $x_1$, and the numerator is the absolute difference of the residuals of $f_P$ and $f_Q$ with a given model $h$ at $x_1$. $\eta$ is the largest ratio between the two absolute differences. A small $\eta$  means even if $f_P$ and $f_Q$ differ significantly, $h$ results in similar prediction residuals on $f_P$ and $f_Q$. When $x_1\neq x_2$, $\eta$ is the largest ratio of the two absolute differences at two positions, $x_1$ and $x_2$, so a small $\eta$ means that $h$ can lead to similar residuals when $f_P(x_1)$ and $f_Q(x_2)$ differ. The expanded definition above includes both $x_1=x_2$ and $x_1\neq x_2$ conditions and Figure \ref{fig: eta visuallization} (a) and (b) present the two conditions, respectively. Intuitively, the residual difference caused by concept shift will be constrained by $\eta$.

\newpage
\section{Distribution estimation}\label{appendix: Distribution Estimation}
\subsection{Kernel density estimation}
$\hat{P}_X$ and $\hat{D}_X^{(i)}$ for $i=1,...,k$ are obtained by kernel density estimation (KDE), and based on the estimated distributions we calculate the likelihood ratio.

In our experiments, we applied the Gaussian kernel, which is a positive function of $x\in\mathcal{X}\subseteq \mathbb{R}^d$ given by
\begin{equation}
    \text{K}(x,b)=\frac{1}{(\sqrt{2\pi}b)^d}e^{-\frac{\lVert x \rVert^2}{2b^2}},
\end{equation}
where $\lVert \cdot \rVert$ is Euclidean distance and $b$ is bandwidth. Given this kernel form, the estimated probability density, denoted by $\hat{p}$, at a position $x_a$ within a group of points $x_1,...,x_n$ is
\begin{equation}
    \hat{p}(x_a,\text{K})=\sum\nolimits_{i=1}^n\text{K}(x_a-x_i,b).
\end{equation}

To find the optimized bandwidth value of $\hat{P}_X$ and $\hat{D}_X^{(i)}$ for $i=1,...,k$ on each dataset, we applied the grid search method with a bandwidth pool using scikit-learn package~\citep{pedregosa2011scikit}. With the approximated marginal distribution densities, we can calculate the likelihood ratio to implement the weighting technique proposed by~\cite{tibshirani2019conformal}.
\subsection{Point-wise distribution estimation}
$\hat{D}_V^{(i)}$ and $\hat{D}_{V,s_P}^{(i)}$ for $i=1,...,k$ are estimated as discontinuous, point-wise distributions to ensure differentiability during training. Specifically, as $\hat{D}_V^{(i)}$ and $\hat{D}_{V,s_P}^{(i)}$ are conformal score distributions on real number set $\mathbb{R}$, $W(\hat{D}_V^{(i)},\hat{D}_{V,s_P}^{(i)})$ is equal to area between their CDFs, as Eq.~(\ref{eq: Wasserstein on real number set}) shows. Hence, our focus is on estimating the CDFs of $\hat{D}_V^{(i)}$ and $\hat{D}_{V,s_P}^{(i)}$ for $i=1,...,k$.

For the details of point-wise distribution estimation, consider we have a $x_1,...,x_n$ drawn from a probability measure $\mu$ in space $\mathcal{X}\subseteq\mathbb{R}$, so the approximated CDF of $\mu$ is given by
\begin{equation}\label{eq: point-wise CDF}
        {F}_{\hat{\mu}}(x)=\frac{1}{n}\sum\nolimits_{j=1}^n\delta_{x_i}\mathbbm{1}_{x_i<x},
\end{equation}
where $\mathbbm{1}$ is the indicator function and $\delta_{x_i}$ represents the point mass at $x_i$ (i.e., the distribution placing all mass at the value $x_i$). In other words, Eq.~(\ref{eq: point-wise CDF}) counts the partition of samples that are smaller than $x$. This point-wise estimation ensures that the Wasserstein-1 distance between the estimated distributions is differentiable.

\section{Supplementary experimental insights}\label{appendix: Experiment}
\subsection{Datasets}\label{appendix_sub: Dataset}
The airfoil self-noise dataset from the UCI Machine Learning Repository~\citep{misc_airfoil_self-noise_291} was intentionally modified to introduce covariate shift and concept shift among them. It includes 1503 instances. The target variable is the scaled sound pressure level of NASA airfoils, and there are 5 features: log frequency, angle of attack, chord length, free-stream velocity, and log displacement thickness of the suction side. To introduce covariate shift, we divided the original dataset into three subsets based on the 33\% and 66\% quantiles of the first dimension feature, log frequency, and partially shuffled them. Therefore, $k=3$ for this dataset. We further introduced concept shifts among the three subsets by modifying target values. With $\xi$ following a normal distribution $N(0,10)$, for $y$ in the first set, $y+=y/1000*\xi$; for $y$ in the second set, $y+=y/\xi$; for $y$ in the third set, $y+=\xi$. With the modified data, we conducted sampling trials to generate 10 randomly sampled datasets.

The Seattle-loop dataset~\cite{cui2019traffic}, as well as the PeMSD4 and PeMSED8 datasets~\cite{guo2019attention}, consist of sensor-observed traffic volume and speed data gathered in Seattle, San Francisco, and San Bernardino, respectively. The data was collected at 5-minute intervals. Our goal for each dataset is to forecast the traffic speed of a specific interested local road segment in the next time step by utilizing the current traffic speed and volume data from both the local segment and its neighboring segments. Before sampling, we selected 10 segments of interest for each dataset randomly, setting $k=10$ for them. There are natural joint distribution shifts present among these segments because of the varying local traffic patterns.

The U.S.-States and Japan-Prefectures datasets~\cite{deng2020cola} contain data on the number of patients infected with influenza-like illness (ILI) reported by the U.S. Department of Health and Human Services, Center for Disease Control and Prevention (CDC), and the Japan Infectious Diseases Weekly Report, respectively. The data in each dataset is structured based on the collection region. Our objective is to utilize the regional predictive features, including population, the increase in the number of infected patients observed in the current week, and the annual cumulative total of infections, to forecast the rise in infections for the following week in the corresponding region. We also randomly selected 10 regions for both datasets, so $k=10$. Due to the diverse regional epidemiological conditions, there are inherent joint distribution shifts among these regions.

For each dataset, we began by sampling $\mathcal{S}^{(i)}_{XY}$ from each subset $i$, for $i=1,...,k$, without replacement. After this step, we allocated the remaining elements within each subset for calibration and testing purposes. The parts intended for calibration across all subsets were then unified to form $\mathcal{S}^{P}_{XY}$. Lastly, to create diverse testing scenarios, we generated multiple test sets by randomly mixing the parts designated for testing from each subset with replacement. For each dataset, we conducted the sampling trial for 10 times, and calculated the mean and standard deviation of the results from these trials, as shown in Figure~\ref{fig: comparison in Wasserstein distance}, Figure~\ref{fig: Coverage and Size 02}, and Figure~\ref{fig: Pareto front}. For efficiency, all CP methods were conducted as split conformal prediction. 

\begin{figure*}[h]
\centering
\captionsetup{singlelinecheck = false, justification=justified}
  \includegraphics[scale=0.25]{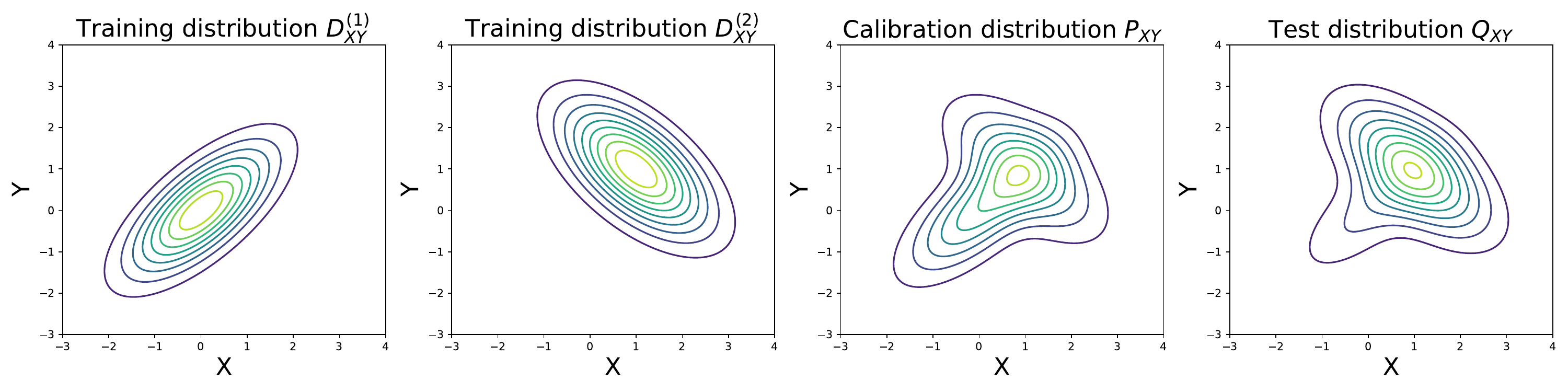}
  \caption{\footnotesize
  \textbf{Calibration and test samples are not exchangeable as they are from different distributions.}}
  \vspace{0pt}
  \label{fig: Data nonexchangeability} 
\end{figure*}

We introduce a toy example to further illustrate that exchangeability does not hold. Consider we have two training distributions: 
\begin{equation*}
    D_{XY}^{(1)}=N\left([0, 0],\begin{bmatrix} 1 & 0.7 \\ 0.7 & 1 \end{bmatrix}\right); D_{XY}^{(2)}=N\left([1, 1],\begin{bmatrix} 1 & -0.6 \\ -0.6 & 1 \end{bmatrix}\right).
\end{equation*}
A calibration distribution is a mixture of these two training distributions with known weights, such as a uniformly weighted mixture ($w_1=w_2=0.5$). A test distribution is a mixture of \( D_{XY}^{(1)} \) and \( D_{XY}^{(2)} \) with unknown random weights. To visualize the non-exchangeability in Figure~\ref{fig: Data nonexchangeability}, we assume the unknown test distribution has weights of \(0.2\) for \( D_{XY}^{(1)} \) and \(0.8\) for \( D_{XY}^{(2)} \).

\subsection{Spearman's coefficient}\label{appendix sub: Coeff}
We first provide the definition of Pearson coefficient.
\begin{definition}[Pearson coefficient]
  With $n$ pairs of samples, $(x_i,y_i)$ for $i=1,...,n$, of two random variables $X$ and $Y$, Pearson coefficient, $r_p$, is calculated as the covariance of the samples divided by the product of their standard deviations. Formally, it is given by 
    \begin{equation}
    \label{eq: pearson coefficient}
         r_p = \frac{\sum\nolimits_{i=1}^n (x_i - \overline{x})(y_i - \overline{y})}{\sqrt{\sum\nolimits_{i=1}^n (x_i - \overline{x})^2}\sqrt{\sum\nolimits_{i=1}^n (y_i - \overline{y})^2}},
    \end{equation}
    where $\overline{x}$ and $\overline{y}$ are the means of the samples of $X$ and $Y$, respectively.
\end{definition}
\newpage
Based on Pearson coefficient, the definition of Spearman's coefficient is given as follows.
\begin{definition}[Spearman's coefficient] With $n$ pairs of samples, $(x_i,y_i)$ for $i=1,...,n$, of two random variables $X$ and $Y$, letting $r(\cdot)$ be the rank function (i.e., $r(x_1)=3$ indicates that $x_1$ is the third largest sample among $x_1,...,x_n$), Spearman's coefficient, $r_s$, is defined as the Pearson coefficient between the ranked samples:
\begin{equation}
    r_s=\frac{\sum\nolimits_{i=1}^n \left(r(x_i) - r(\overline{x})\right)\left(r(y_i) - r(\overline{y})\right)}{\sqrt{\sum\nolimits_{i=1}^n \left(r(x_i) - r(\overline{x})\right)^2}\sqrt{\sum\nolimits_{i=1}^n \left(r(y_i) - r(\overline{y})\right)^2}},
\end{equation}
    where $\overline{x}$ and $\overline{y}$ are the means of the samples of $X$ and $Y$, respectively.
\end{definition}\label{def: Spearman coeff}
We calculated Spearman's coefficient between each distance measure and the largest coverage gap in Section~\ref{sec: experiment} to confirm that Wasserstein distance holds the strongest positive correlation compared with other distance measures.

\subsection{Additional experiment results of subsection~\ref{subsec: coverage and size result}}\label{appendix sub: additional result}
\begin{figure*}[p]
\centering
\captionsetup{singlelinecheck = false, justification=justified}
  \includegraphics[scale=0.4]{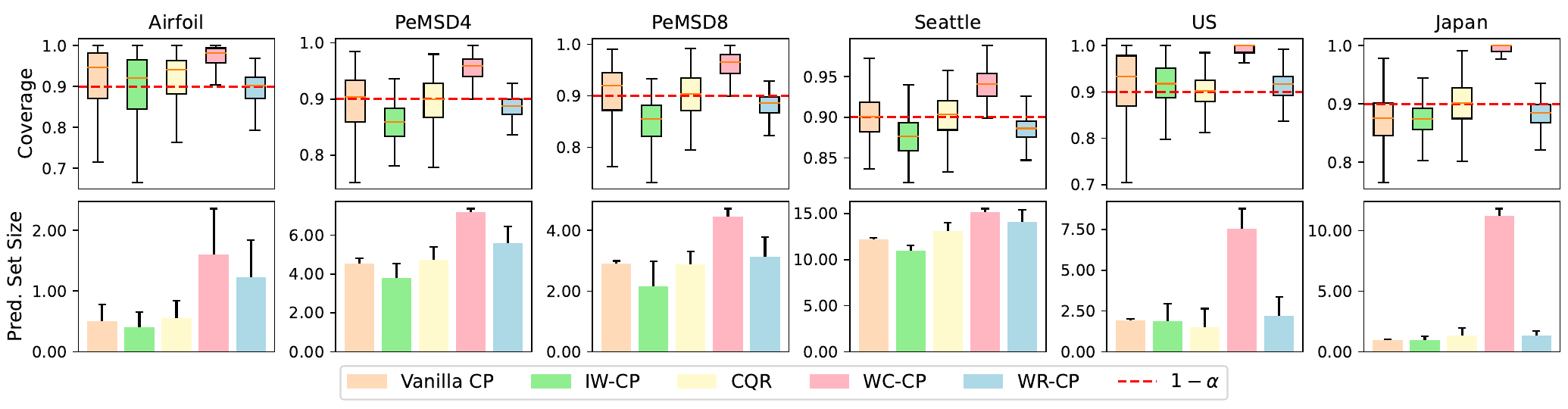}
  \caption{\footnotesize
  \textbf{Coverages and set sizes of WR-CP and baselines with $\alpha=0.1$:} The $\beta$ values for the WR-CP method are 9, 11, 9, 8, 13, and 20, respectively.}
  \vspace{0pt}
  \label{fig: Coverage and Size 01} 
\end{figure*}

\begin{figure*}[p]
\centering
\captionsetup{singlelinecheck = false, justification=justified}
  \includegraphics[scale=0.4]{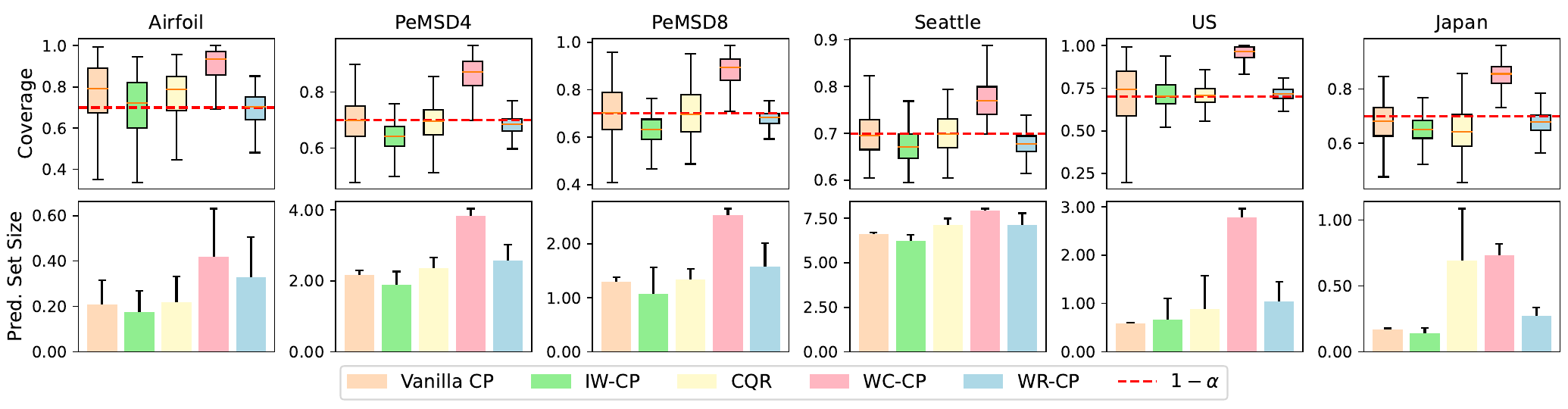}
  \caption{\footnotesize
  \textbf{Coverages and set sizes of WR-CP and baselines with $\alpha=0.3$:} The $\beta$ values for the WR-CP method are 3, 5, 5, 5, 8, and 13, respectively.}
  \vspace{0pt}
  \label{fig: Coverage and Size 03} 
\end{figure*}

\begin{figure*}[p]
\centering
\captionsetup{singlelinecheck = false, justification=justified}
  \includegraphics[scale=0.4]{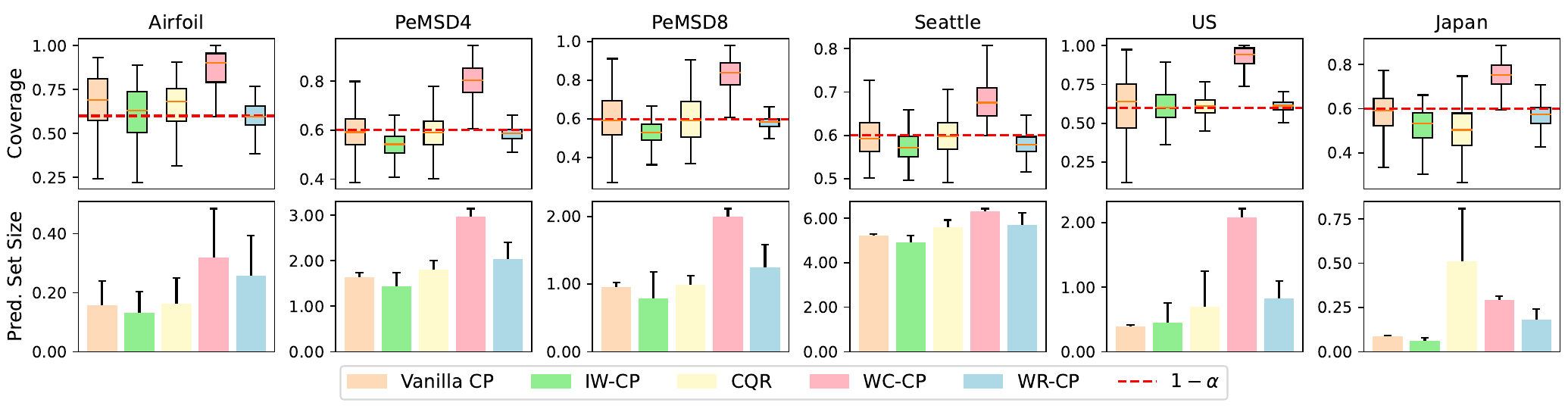}
  \caption{\footnotesize
  \textbf{Coverages and set sizes of WR-CP and baselines with $\alpha=0.4$:} The $\beta$ values for the WR-CP method are 3, 5, 5, 5, 8, and 13, respectively.}
  \vspace{0pt}
  \label{fig: Coverage and Size 04} 
\end{figure*}

\begin{figure*}[p]
\centering
\captionsetup{singlelinecheck = false, justification=justified}
  \includegraphics[scale=0.4]{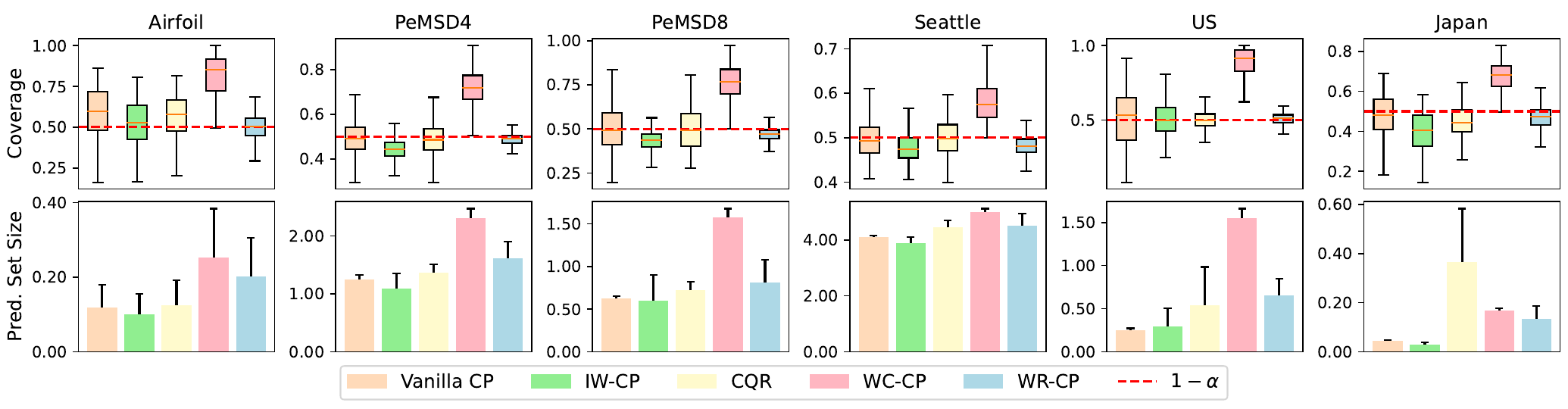}
  \caption{\footnotesize
  \textbf{Coverages and set sizes of WR-CP and baselines with $\alpha=0.5$:} The $\beta$ values for the WR-CP method are 3, 5, 3, 5, 8, and 13, respectively.}
  \vspace{0pt}
  \label{fig: Coverage and Size 05}  
\end{figure*}

\begin{figure*}[p]
\centering
\captionsetup{singlelinecheck = false, justification=justified}
  \includegraphics[scale=0.4]{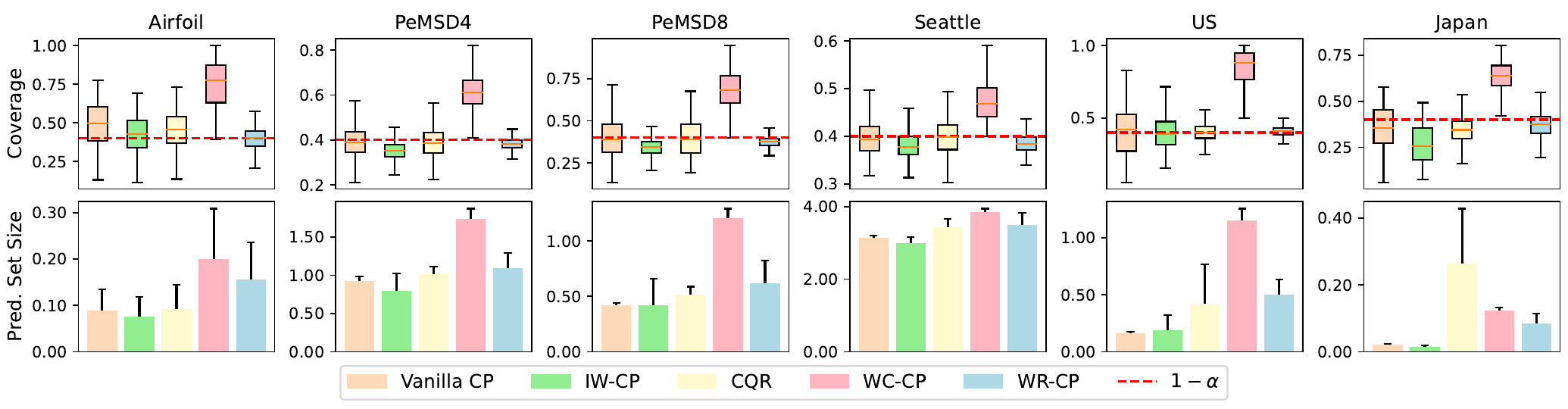}
  \caption{\footnotesize
  \textbf{Coverages and set sizes of WR-CP and baselines with $\alpha=0.6$:} The $\beta$ values for the WR-CP method are 3, 5, 3, 5, 8, and 13, respectively.}
  \vspace{0pt}
  \label{fig: Coverage and Size 06}  
\end{figure*}

\begin{figure*}[p]
\centering
\captionsetup{singlelinecheck = false, justification=justified}
  \includegraphics[scale=0.4]{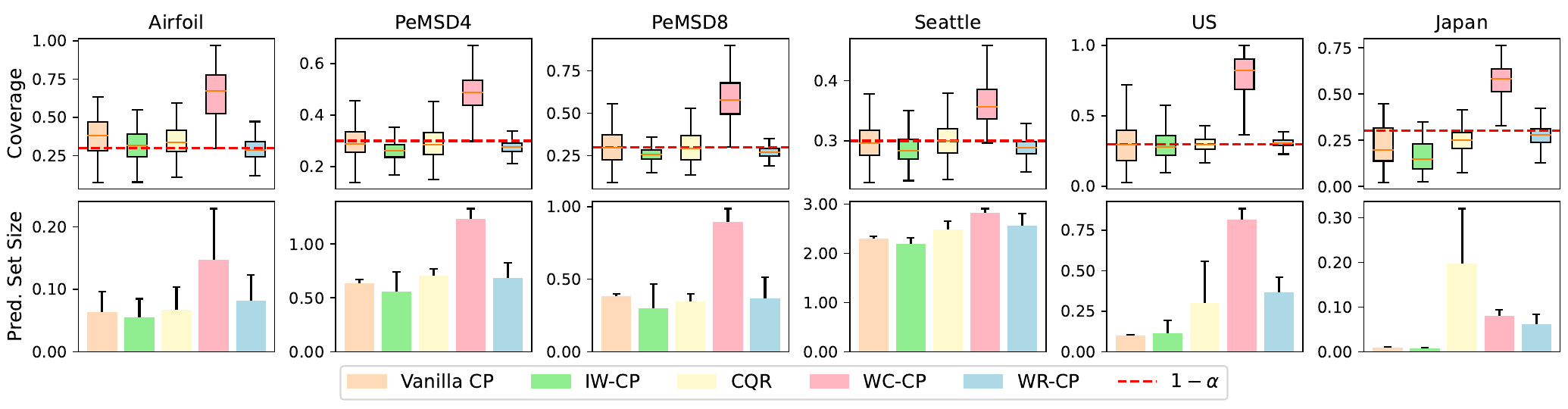}
  \caption{\footnotesize
  \textbf{Coverages and set sizes of WR-CP and baselines with $\alpha=0.7$:} The $\beta$ values for the WR-CP method are 2, 2, 2, 5, 8, and 10, respectively.}
  \vspace{0pt}
  \label{fig: Coverage and Size 07}  
\end{figure*}

\begin{figure*}[p]
\centering
\captionsetup{singlelinecheck = false, justification=justified}
  \includegraphics[scale=0.4]{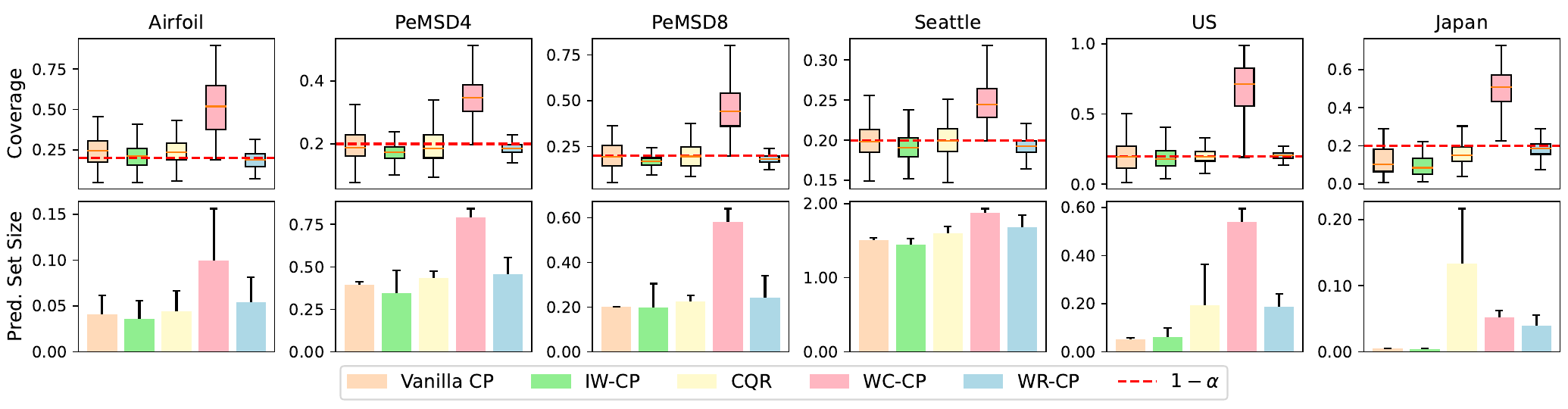}
  \caption{\footnotesize
  \textbf{Coverages and set sizes of WR-CP and baselines with $\alpha=0.8$:} The $\beta$ values for the WR-CP method are 2, 2, 2, 5, 5, and 10, respectively.}
  \vspace{0pt}
  \label{fig: Coverage and Size 08}  
\end{figure*}

\begin{figure*}[p]
\centering
\captionsetup{singlelinecheck = false, justification=justified}
  \includegraphics[scale=0.4]{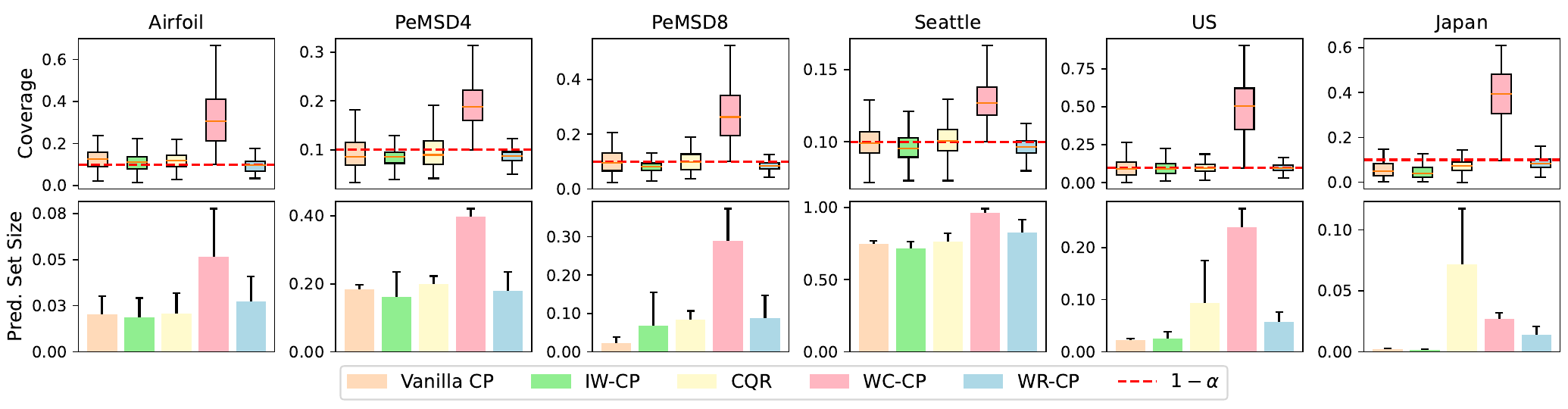}
  \caption{\footnotesize
  \textbf{Coverages and set sizes of WR-CP and baselines with $\alpha=0.9$:} The $\beta$ values for the WR-CP method are 2, 1, 1, 5, 2, and 6, respectively.}
  \vspace{0pt}
  \label{fig: Coverage and Size 09}  
\end{figure*}

In addition to the results shown in Figure~\ref{fig: Coverage and Size 02}, we present further experimental findings from Subsection~\ref{subsec: coverage and size result} with $\alpha$ values of 0.1, 0.3, 0.4, 0.5, 0.6, 0.7, 0.8, and 0.9 on Figure~\ref{fig: Coverage and Size 01},~\ref{fig: Coverage and Size 03},~\ref{fig: Coverage and Size 04},~\ref{fig: Coverage and Size 05},~\ref{fig: Coverage and Size 06},~\ref{fig: Coverage and Size 07},~\ref{fig: Coverage and Size 08}, and~\ref{fig: Coverage and Size 09}, respectively. Clearly, WR-CP demonstrates the ability to generate more tightly concentrated coverages near $1-\alpha$ compared to vanilla CP and IW-CP. Additionally, it yields smaller prediction set sizes than the state-of-the-art method WC-CP. These figures also reveal a trend where as the $\alpha$ value increases, WR-CP requires a smaller $\beta$ to achieve acceptable coverages around $1-\alpha$, so the prediction set sizes produced by WR-CP are closer with those of vanilla CP and IW-CP, as evidenced by the results on the PeMSD4 in Figure~\ref{fig: Coverage and Size 01} and Figure~\ref{fig: Coverage and Size 09}. This phenomenon could be attributed to the trade-off between conformal prediction accuracy and efficiency under joint distribution shift. The Wasserstein regularization term in Eq.~(\ref{eq: objective function in population}) tends to prioritize aligning smaller conformal scores initially, as it reduces the Wasserstein penalty with a lesser increase in the empirical risk minimization term. Hence, as the hyperparameter $\beta$ increases, the model gradually aligns larger conformal scores from two different distributions, which will adversely impact the risk-driven term more. When considering a higher $\alpha$ value, the focus is on ensuring that the coverages on test data are close to the smaller $1-\alpha$, indicating the importance of aligning small conformal scores.  Consequently, a high $\beta$ value is not necessary in this case, leading to smaller prediction set sizes being achieved.

\subsection{Experiment setups in ablation study}\label{appendix sub: beta values}
To visualize a comprehensive and evenly-distributed set of optimal solutions on Pareto fronts, we utilized WR-CP with varying values of $\beta$ to produce the results depicted in Figure~\ref{fig: Pareto front}. As mentioned in Section~\ref{sec: algorithm}, it is worth noting that when $\beta=0$, WR-CP reverts to IW-CP. The selected $\beta$ values for the results of Figure~\ref{fig: Pareto front} are shown in Table~\ref{table: beta values}.
\begin{table}[h]
\small
\centering
\caption{$\beta$ values of WR-CP in ablation study}
\def\arraystretch{1.2}
\begin{tabular}{|c|l|}
\hline
Dataset & \multicolumn{1}{c|}{$\beta$ values} \\ \hline
Airfoil & 1, 1.5, 2, 2.5, 3, 3.5, 4.5, 6, 8, 9, 13, 20. \\ \hline
PeMSD4 & 1, 1.5, 2, 2.5, 3, 5, 7, 9, 11, 15, 20. \\ \hline
PeMSD8 & 1, 1.5, 2, 2.5, 3, 4, 5, 7, 9, 17. \\ \hline
Seattle & \multicolumn{1}{c|}{1, 2, 3, 4, 4.5, 5, 5.5, 6, 7, 8, 10, 13, 15, 20.} \\ \hline
U.S. & 1, 1.5, 2, 2.5, 3, 5, 6, 8, 13. \\ \hline
Japan & 1, 2, 3, 4, 6, 8, 10, 13, 20. \\ \hline
\end{tabular}
\label{table: beta values}
\end{table}
\newpage
\section{Prediction efficiency with coverage guarantee}\label{appendix: prediction efficiency with coverage guarantee}
Although Wasserstein-regularized loss in Eq.~(\ref{eq: objective function in population}) offers a controllable trade-off with significantly improved prediction efficiency and a mild coverage loss, it is worth investigating if this efficiency can be achieved with a coverage guarantee. In this section, we first derive a coverage lower bound of WR-CP via the multi-source setup in Appendix~\ref{appendix sub: guarantee from multi-source}. Then, we show that the combination of WC-CP and the Wasserstein-regularized loss can not achieve small prediction sets with ensured coverage in 
Appendix~\ref{appendix sub: hybrid WR-WC}.

\subsection{Coverage guarantee from multi-source setup}\label{appendix sub: guarantee from multi-source}
Under the setup of multi-source conformal prediction, with $\tau$ as the $1-\alpha$ quantile of the weighted calibration conformal score distribution $Q_{V,s_P}$, we can derive the coverage gap upper bound by
\begin{equation}\label{eq: gap mixture bound}
\begin{split}
        |F_{Q_{V,s_P}}(\tau) - F_{Q_{V}}(\tau)| &= \left| \sum_{i=1}^k w_i F_{D_{V,s_P}^{(i)}}(\tau) - \sum_{i=1}^k w_i F_{D^{(i)}_{V}}(\tau) \right|\\&\leq \sum_{i=1}^k w_i |F_{D^{(i)}_{V,s_P}}(\tau) - F_{D^{(i)}_{V}}(\tau)|\\& \leq\sup_{i\in\{1,...,k\}} |F_{D^{(i)}_{V,s_P}}(\tau) - F_{D^{(i)}_{V}}(\tau)|.
\end{split}
\end{equation}
In other words, the coverage gap on a test distribution must be less or equal to the largest gap at $\tau$ among multiple training distributions. Denoting $\alpha_D=\sup_{i\in\{1,...,k\}} |F_{D^{(i)}_{V,s_P}}(\tau) - F_{D^{(i)}_{V}}(\tau)|$, we have a coverage guarantee $\text{Pr}(Y_{n+1}\in X_{n+1})\geq 1-\alpha-\alpha_D$.

The regularization term $\sum_{i=1}^kW(D^{(i)}_{V,s_P}, D^{(i)}_V)$ in Eq.~(\ref{eq: objective function in population}) can minimize $\alpha_D$, and thus making $1-\alpha-\alpha_D$ closer to the desired $1-\alpha$. It is important to highlight that $\alpha_D$ is adaptive to variations in test distribution $Q_V$, as evident from Eq.~(\ref{eq: gap mixture bound}). This adaptivity ensures that the lower bound dynamically adjusts to different $Q_V$. To evaluate the prediction efficiency of WR-CP under this guarantee, we set $\alpha=0.1$ and computed the corresponding $\alpha_D$ for various test distributions. Additionally, we calculated the coverage and prediction set size of WC-CP on each test distribution, using the corresponding guarantee at $1 - \alpha - \alpha_D$ for comparison.

\begin{figure*}[h]
\centering
\captionsetup{singlelinecheck = false, justification=justified}
  \includegraphics[scale=0.4]{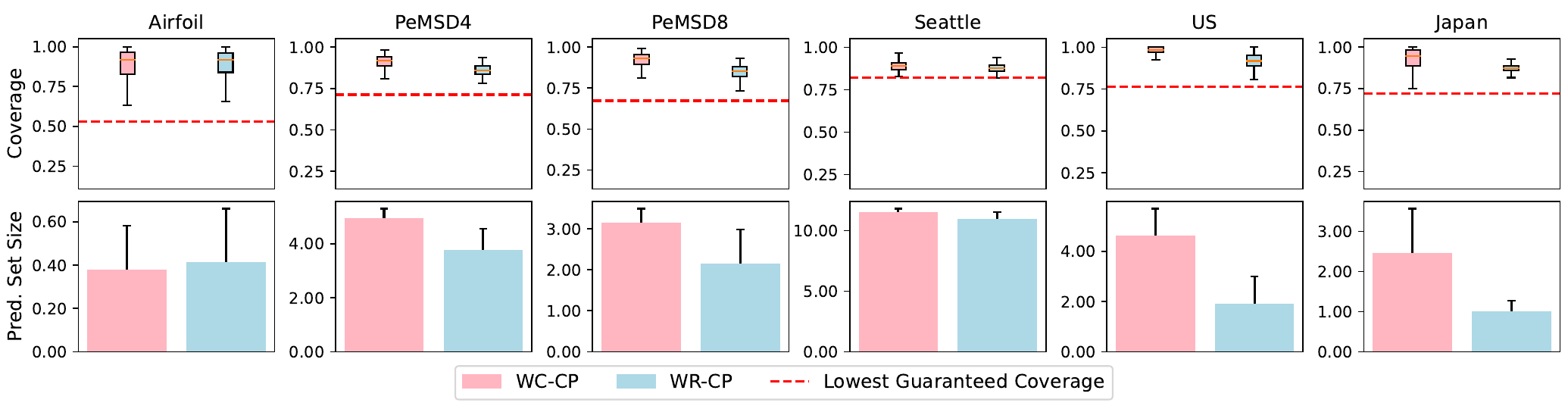}
  \caption{\footnotesize
  \textbf{Coverages and set sizes of WC-CP and WR-CP with coverage guarantee at $1-\alpha-\alpha_D$.}}
  \vspace{0pt}
  \label{fig: Guaranteed by mixture 01}  
\end{figure*}

The experiment results are depicted in Figure~\ref{fig: Guaranteed by mixture 01}, demonstrating improved prediction efficiency on the PeMSD4, PeMSD8, U.S.-States, and Japan-Prefectures datasets. However, the efficiency remains almost unchanged on the Seattle-loop dataset and even declines on the airfoil self-noise dataset. This phenomenon can be attributed to the regularization mechanism. While WR-CP enhances prediction efficiency by leveraging the calibration distribution to generate prediction sets, regularization inevitably increases prediction residuals, leading to larger prediction sets. These two opposing effects can interact differently depending on the dataset characteristics. When the efficiency gains outweigh the drawbacks of regularization, we observe reduced prediction set size. Conversely, in datasets like the Seattle-loop and airfoil self-noise, the benefits of regularization are outweighed by the increased prediction residuals, resulting in unchanged or diminished efficiency. The averaged prediction set size reduction across the six datasets is $26.9\%$.

\subsection{Poor compatibility between Wasserstein-regularized loss and WC-CP}~\label{appendix sub: hybrid WR-WC}
\begin{figure*}[h]
\centering
\captionsetup{singlelinecheck = false, justification=justified}
  \includegraphics[scale=0.4]{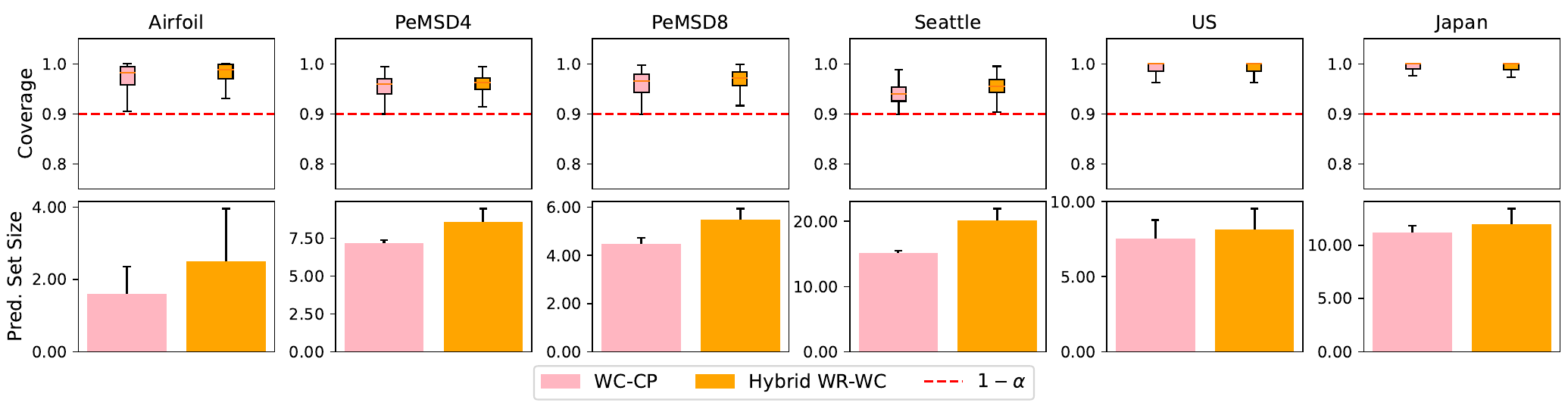}
  \caption{\footnotesize
  \textbf{Coverages and set sizes of WC-CP and Hybrid WC-WR with coverage guarantee $1-\alpha=0.9$.}}
  \vspace{0pt}
  \label{fig: Guaranteed by Hybrid 01}  
\end{figure*}

Since the WC-CP is a conservative \textit{post-hoc} uncertainty quantification method but the proposed regularized loss in Eq.~(\ref{eq: objective function in population}) is applied during \textit{training}, one may consider applying WC-CP upon the model trained by the regularized loss to obtain guaranteed coverage. However, WC-CP and the model are not suitable for complementing each other. While regularization enhances the reliability of calibration distributions, the worst-case approach depends exclusively on the upper bound of $1-\alpha$ test conformal score quantile, rendering it unable to benefit from regularization. In contrast, the WC-CP may result in larger prediction sets under this condition, as the regularization inevitably increases the prediction residuals, which in turn increases the upper bound of the test conformal score quantile. Experiment results in Figure~\ref{fig: Guaranteed by Hybrid 01} demonstrate the analysis, where WC-CP is the worst-case method based on a residual-driven model (same as the WC-CP method in Section~\ref{subsec: coverage and size result}), and Hybrid WC-WR represents applying WC-CP to a model trained by Eq.~(\ref{eq: objective function in population}).

\section{Limitations}\label{appendix: limitation}
\subsection{Susceptibility to density estimation errors}
Given that Wasserstein regularization relies on importance-weighted conformal scores, its performance is greatly influenced by the accuracy of the estimated likelihood ratio obtained through KDE. Inaccurate estimation can significantly impact the effectiveness of WR-CP. For instance, in Figure~\ref{fig: Coverage and Size 02}, WR-CP yields larger prediction set sizes with less concentrated coverages on the airfoil self-noise dataset compared to other datasets. This can be attributed to the airfoil self-noise dataset having the highest feature dimension (5) and the smallest size of the sampled $\mathcal{S}^P_{XY}$ (500). These challenges in KDE lead to suboptimal performance of WR-CP on the airfoil self-noise dataset when compared to its performance on others.

The main reason for KDE error is numerical instability, which can arise from several factors. A poor choice of kernel is a critical contributor; for instance, kernels with sharp edges or discontinuities, such as rectangular or triangular kernels, can result in jagged density estimates and amplify errors near boundaries. Fat-tailed kernels, such as the Cauchy kernel, may assign excessive weight to distant data points, leading to inaccuracies in density estimates and numerical precision challenges. Additionally, the lack of feature normalization can exacerbate the effects of extreme values, skewing the density estimation process and reducing computational stability. Lastly, inappropriate bandwidth selection, either too small (overfitting) or too large (underfitting), can disrupt the balance between bias and variance, further contributing to instability in the estimation.

In our work, we first adopted the Gaussian kernel, valued for its smoothness and numerical stability. To mitigate the influence of extreme values, we applied feature normalization, ensuring a more stable density estimation process. Additionally, we conducted a comprehensive grid search to fine-tune the bandwidth, achieving an optimal balance between bias and variance for robust and accurate results. The bandwidth candidates were selected from a logarithmically spaced range between \(10^{-2}\) and \(10^{0.5}\), consisting of 20 evenly distributed values on a logarithmic scale. 

\subsection{Computational challenges in kde}

We applied a grid search approach to identify the optimal bandwidth for KDE, which ensures an effective balance between bias and variance in density estimation. However, this method often involves extensive computational effort, particularly when working with high-dimensional datasets, as it requires repeated calculations over a range of bandwidth values. To address this challenge, Bernacchia–Pigolotti KDE~\citep{bernacchia2011self} introduces an innovative framework that combines a Fourier-based filter with a systematic approach for simultaneously determining both the kernel shape and bandwidth. This method not only reduces subjectivity in kernel selection but also offers a more efficient computational pathway. Building on this foundation, FastKDE~\citep{o2016fast} adapts and extends the Bernacchia–Pigolotti approach for high-dimensional scenarios, incorporating optimizations that significantly improve computational speed and scalability. These advancements represent promising directions for mitigating the computational overhead in our own work, where similar strategies could be leveraged to streamline the bandwidth selection process and enhance the overall efficiency of KDE in complex datasets.

\subsection{Other choices of the calibration distribution}
In the experiments conducted in Section~\ref{sec: experiment}, we specifically examine the scenario where the calibration data follows a mixture distribution of $D_{XY}^{(i)}$ for $i=1,...,k$ with equal weights. However, this may not always be the case in real-world situations. Given that the calibration distribution plays a crucial role in determining the difficulty of minimizing Eq.~(\ref{eq: objective function in population}) during training, it is valuable to investigate the performance of WR-CP with a calibration distribution different from a mixture of training distributions.

\end{document}